\documentclass[]{interact}

\usepackage{epstopdf}

\makeatother

\usepackage{hyperref}
\hypersetup{
    colorlinks=true,   
    linkcolor=black,   
    citecolor=black,   
    urlcolor=black     
}

\usepackage{amsmath}
\usepackage{amssymb}
\usepackage{mathtools}
\usepackage{amsthm}
\usepackage{algorithm}
\usepackage[noend]{algpseudocode}

\usepackage[printonlyused]{acronym}

\usepackage{graphicx}
\usepackage{multirow}
\usepackage{makecell}
\usepackage{tabularx} 
\usepackage{colortbl}
\usepackage{placeins}

\usepackage{pifont}

\newcommand{\eg}{e.g., }
\newcommand{\ie}{i.e., }
\newcommand{\cf}{cf. }

\newcommand{\sota}{state-of-the-art }

\DeclareMathOperator*{\argmin}{arg\,min}

\usepackage{acronym}
\acrodef{ASR}{Attack Success Rate}
\acrodef{BGI}{Batch recovery via Gradient Inversion}
\acrodef{BN}{Batch Normalization}
\acrodef{CB}{Client Benefit}
\acrodef{CDIA}{Combined Defenses Inversion Attack}
\acrodef{CFL}{Clustered Federated Learning}
\acrodef{CH}{Classification Head}
\acrodef{CNN}{Convolutional Neural Network}
\acrodef{CPL}{Client Privacy Leakage}
\acrodef{CVB}{Convolutional Variational Bottleneck}
\acrodef{DGC}{Deep Gradient Compression}
\acrodef{DIA}{Dropout Inversion Attack}
\acrodef{DL}{Deep Learning}
\acrodef{DLG}{Deep Leakage from Gradients}
\acrodef{DP}{Differential Privacy}
\acrodef{DPIA}{Differential Privacy Inversion Attack}
\acrodef{DPSGD}{Differentially Private Stochastic Gradient Descent}
\acrodef{EP}{Empirical Privacy}
\acrodef{FE}{Feature Extractor}
\acrodef{FedAvg}{Federated Averaging}
\acrodef{FFP}{Feature-base FingerPrinting}
\acrodef{FL}{Federated Learning}
\acrodef{FLT}{Federated Learning with Taskonomy}
\acrodef{FTL}{Federated Transfer Learning}
\acrodef{GAN}{Generative Adversarial Network}
\acrodef{GDF}{Gradient Distance Function}
\acrodef{GELU}{Gaussian Error Linear Unit}
\acrodef{GI}{Gradient Inversion}
\acrodef{GN}{Group Normalization}
\acrodef{GP}{Gradient Pruning}
\acrodef{GPIA}{Gradient Pruning Inversion Attack}
\acrodef{HC}{Hierarchical Clustering}
\acrodef{HE}{Homomorphic Encryption}
\acrodef{HFL}{Horizontal Federated Learning}
\acrodef{IID}{Independent and Identically Distributed}
\acrodef{iDLG}{improved Deep Leakage from Gradients}
\acrodef{IFCA}{Iterative Federated Clustering Algorithm}
\acrodef{IG}{Inverting Gradients}
\acrodef{KLD}{Kullbach-Leibler Divergence}
\acrodef{LPIPS}{Learned Perceptual Image Patch Similarity}
\acrodef{LRM}{Label Reconstruction Method}
\acrodef{MD}{Mask Distance}
\acrodef{MI}{Model Inversion}
\acrodef{ML}{Machine Learning}
\acrodef{MLP}{Multi Layer Perceptron}
\acrodef{MMD}{Maximum Mean Discrepancy}
\acrodef{MSE}{Mean Squared Error}
\acrodef{NN}{Neural Network}
\acrodef{PACFL}{Principal Angles Analysis for Clustered Federated Learning}
\acrodef{PCA}{Principal Component Analysis}
\acrodef{PFL}{Personalized Federated Learning}
\acrodef{PIG}{Partial Inverting Gradients}
\acrodef{PM}{Privacy Module}
\acrodef{PRECODE}{PRivacy EnhanCing mODulE}
\acrodef{PSNR}{Peak Signal to Noise Ratio}
\acrodef{SAPAG}{Self-Adaptive Privacy Attack from Gradients}
\acrodef{SGD}{Stochastic Gradient Descent}
\acrodef{SMPC}{Secure Multi-Party Computation}
\acrodef{SSIM}{Structural Similarity}
\acrodef{SVD}{Singular Value Decomposition}
\acrodef{t-SNE}{t-Distributed Stochastic Neighbor Embedding}
\acrodef{TRD}{Throughput Rate Distance}
\acrodef{TV}{Total Variation}
\acrodef{VB}{Variational Bottleneck}
\acrodef{VBIA}{Variational Bottleneck Inversion Attack}
\acrodef{VFL}{Vertical Federated Learning}
\acrodef{ViT}{Vision Transformer}
\acrodef{WIIG}{Well-Informed Inverting Gradients}

\newcommand{\vLayer}{\ell}
\newcommand{\vlayerindex}{l}
\newcommand{\vLayerindex}{L}

\newcommand{\vrunningindexi}{i}

\newcommand{\vthreshclip}{\tau_{\text{clip}}}

\newcommand{\vprune}{p_{\text{prune}}}

\newcommand{\vdrop}{p_{\text{dr}}}

\newcommand{\vPrior}{\mathcal{R}}

\newcommand{\vthreshASR}{\tau_{\text{ASR}}}

\newcommand{\vFeaturevector}{z}

\newcommand{\vVBeps}{\varepsilon_{\text{VB}}}
\newcommand{\vVBbottlerep}{\beta_{\text{VB}}}

\newcommand{\cmark}{\ding{51}}%
\newcommand{\xmark}{\ding{55}}%

\usepackage{todonotes}

\usepackage[style=authoryear, maxcitenames=1, backend=bibtex]{biblatex}
\renewcommand{\paragraph}[1]{%
    \vspace{0.5\baselineskip}
    \noindent\textbf{#1}\\
}

\begin{document}

\title{Combining Stochastic Defenses to Resist Gradient Inversion: An Ablation Study}

\author{
\name{Daniel Scheliga\textsuperscript{a}, Patrick Mäder\textsuperscript{a,b} and Marco Seeland\textsuperscript{a}\thanks{Corresponding author: Marco Seeland. Email: marco.seeland@tu-ilmenau.de}}
\affil{\textsuperscript{a}Department of Computer Science and Automation, Data-intensive Systems and Visualization Group (dAI.SY), Technische Universität Ilmenau, Ilmenau, Germany}
\affil{\textsuperscript{b}Faculty of Biological Sciences, Friedrich
Schiller University, Jena, Germany}
}
\maketitle

\begin{abstract}
\acf{GI} attacks are a ubiquitous threat in \acf{FL} as they exploit gradient leakage to reconstruct supposedly private training data.
Common defense mechanisms such as \acf{DP} or stochastic \acfp{PM} introduce randomness during gradient computation to prevent such attacks.
However, we pose that if an attacker effectively mimics a client's stochastic gradient computation, the attacker can circumvent the defense and reconstruct clients' private training data.
This paper introduces several targeted \ac{GI} attacks that leverage this principle to bypass common defense mechanisms.
As a result, we demonstrate that no individual defense provides sufficient privacy protection.
To address this issue, we propose to combine multiple defenses.
We conduct an extensive ablation study to evaluate the influence of various combinations of defenses on privacy protection and model utility.
We observe that only the combination of \ac{DP} and a stochastic \ac{PM} was sufficient to decrease the \acf{ASR} from 100\% to 0\%, thus preserving privacy.
Moreover, we found that this combination of defenses consistently achieves the best trade-off between privacy and model utility.
\end{abstract}


\section{Introduction}
\label{sec:intro}
\acf{FL} leverages distributed data to collaboratively improve the utility of neural networks. 
Multiple clients exchange local training gradients to collaboratively train a common global model.
This eliminates the need for centrally aggregated or shared data.
Because training data remains local to each participating client, such collaborative learning systems aim to systematically mitigate privacy risks \parencite{mcmahan2017communication, kairouz2021advances}.
However, potentially sensitive information can be reconstructed from the exchanged gradient information, thereby compromising client privacy.
Iterative \acf{GI} attacks are particularly advanced in this context~\parencite{zhu2019deep, zhao2020idlg, wei2020framework, geiping2020inverting, yin2021see}. 
These attacks optimize initially random dummy data to minimize a distance function between dummy gradients and the attacked client’s gradients.
As a result, they can achieve close to perfect reconstructions of the clients' training data.

The de-facto standard defense against such privacy leaks is to perturb the exchanged gradients by applying \acf{DP} or \ac{GP}~\parencite{bonawitz2017practical, jayaraman2019evaluating, zhu2019deep, sattler2019robust, jin2020stochastic, wei2021gradient, ponomareva2023dp}.
However, gradient perturbation results in an inherent trade-off between model utility and privacy~\parencite{dwork2014algorithmic, jayaraman2019evaluating, scheliga2022precode, huang2021evaluating, ponomareva2023dp}.
To avoid this trade-off, \acfp{PM} such as \ac{PRECODE}~\parencite{scheliga2022precode} and \acp{CVB}~\parencite{scheliga2024privacy} have been proposed.
Related work has found that Dropout has similar privacy preserving effects due to its stochasticity during training~\parencite{scheliga2023dropout}. 

\begin{figure}[t!]
    \centering
    \includegraphics[width=0.99\linewidth]{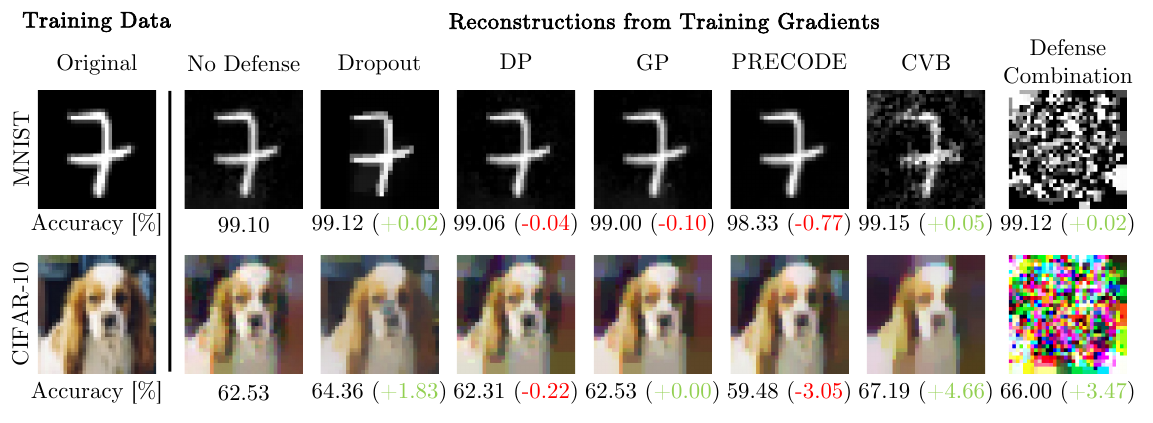}
    \caption{
    \textbf{Visual summary of this paper:}
    Neural networks are trained on MNIST, CIFAR-10, and four other datasets in a \ac{FL} scenario.
    If defenses are applied on their own, they can be bypassed by targeted attacks or require such high gradient perturbation rates, that the resulting model would suffer from severe losses in model utility.
    However, applying a combination of defense mechanisms can prevent leakage from targeted \ac{GI} attacks and even improve model utility compared to an unprotected baseline model.
    }
    \label{fig:teaser}
\end{figure}

To test the limits of these commonly applied defense mechanisms, we adopt the perspective of an attacker. 
Recent work on a \ac{DIA} shows that an attacker can bypass the privacy preserving effects of Dropout by approximating the Dropout masks used by the clients during local training~\parencite{scheliga2023dropout}.
Based on this observation, we argue that training data can be reconstructed even from protected gradients if the attacker sufficiently mimics the clients' stochastic gradient computation processes. 
By applying this attack principle to other defenses, we identify new vulnerabilities in common defense mechanisms in general.
Specifically, we propose targeted \ac{GI} attacks against \ac{DP}~\parencite{abadi2016deep}, \ac{GP}~\parencite{lin2017deep} and two variational modeling based \acp{PM}: \ac{PRECODE}~\parencite{scheliga2022precode} and \ac{CVB}~\parencite{scheliga2024privacy}.
By approximating and mimicking the client's stochastic gradient computation behavior, these targeted attacks significantly improve the \acf{ASR}.
As a result, we observe that an isolated defense mechanism cannot sufficiently protect the data from reconstruction.

To address this privacy issue, we propose to combine multiple defense mechanisms. 
Since each defense mechanism introduces a different type of stochasticity into the training process, the complexity of \ac{GI} attacks increases, making it harder for the attacker to mimic the gradient computation process. 
To rigorously test the effectiveness of various defense combinations, we introduce the \emph{\acf{CDIA}}, a \ac{GI} attack that is aware of the defense mechanisms employed by the client during local training and specifically targets each defense in the attack.
An extensive ablation study of defense combinations shows that combining low-level noise \ac{DP} with a \ac{PM} (\ac{PRECODE} or \ac{CVB}) is essential to protect client privacy and reduce the \ac{ASR} to $0\%$. 
Moreover, this combination often results in the highest model utility compared to other defense combinations and the unprotected baseline model.

\autoref{fig:teaser} presents a visual summary of this paper.
The specific contributions of this paper can be summarized as follows: 
\begin{itemize} 
    \item We investigate four different defense techniques, namely \ac{DP}, \ac{GP}, \ac{PRECODE}, and \ac{CVB}, to analyze differences in the gradient computation process and the resulting gradients between client and attacker.
    \item We introduce three novel \ac{GI} attacks, i.e., \ac{DPIA}, \ac{GPIA}, and \ac{VBIA}, that specifically target these defense mechanisms by approximating and mimicking the client's gradient computation behavior. 
    \item We systematically combine defense techniques and perform an empirical ablation study using our proposed \acf{CDIA}.
\end{itemize}

\section{Related Work}
\subsection{Attacks}
Various studies have demonstrated that \ac{FL} is vulnerable to security and privacy threats~\parencite{mothukuri2021survey, rodriguez2023survey, rao2024privacy, sharma2024review}.
This paper specifically considers \ac{GI} attacks.
This privacy threat is of particular interest, because attackers can achieve almost perfect reconstruction of clients' training data from the gradient information that is exchanged during collaborative training~\parencite{zhu2019deep, geiping2020inverting, wei2020framework, rodriguez2023survey}.
Recent work on \ac{GI} attacks discusses malicious threat models that actively interfere with the training process to create favorable conditions for the \ac{GI} attack~\parencite{fowl2021robbing, fowl2022decepticons, wen2022fishing, boenisch2023curious}.
However, clients could detect such malicious behavior and stop participation in training~\parencite{boenisch2023curious, garov2023hiding}.
Therefore, this work focuses on the honest-but-curious server threat model, which will be discussed in more detail in \autoref{sec:threat_model}.
\ac{GI} attacks can be further categorized into analytical/recursive \ac{GI} attacks and iterative ones.
The authors of~\parencite{aono2017privacy, geiping2020inverting} show that the input to any fully connected layer in a neural network can be analytically reconstructed using the layer's bias values.
Recursive \ac{GI} attacks extend analytical attacks to models that use more than one fully connected layer and get rid of the dependency on the bias term for reconstruction~\parencite{zhu2021r}.
However these attacks are not very successful for larger batch sizes, since only the average of multiple inputs will be recovered.

Iterative \ac{GI} attacks solve an optimization problem to minimize a distance function between dummy gradients and the attacked client gradients.
The dummy data is iteratively adjusted to fit the client gradients.
As a result, the optimized dummy data resembles the client's private training data~\parencite{zhu2019deep, zhao2020idlg, wei2020framework, geiping2020inverting}.
\autoref{sec:prelim_GI} will explain how these attacks work in more detail.
Most research on \ac{GI} attacks proposes some adjustments to some of the attack components to slightly increase the quality of the reconstructed data, stabilize the attack optimization process and/or increase the convergence speed~\parencite{wang2020sapag, jeon2021gradient, yin2021see}.
Recent work also applies generative modeling to increase the fidelity of reconstructions~\parencite{jeon2021gradient, ren2022grnn, xu2022cgir, li2022auditing, zhang2023generative, fang2023gifd}.
However, they often require additional knowledge on the training data distribution of the victim client and auxiliary data.
Furthermore, although the reconstructions do represent the victim clients training data distribution they might be sufficiently different from the actual training data.
For a more detailed overview of the field of \ac{GI}, we refer interested readers to several survey papers \parencite{zhang2022survey, yang2023gradient, li2023survey}.

\subsection{Defenses}
\label{sec:sota_def}
A comprehensive analysis of privacy leakage via \ac{GI} attacks identified relevant parameters that affect the severity of privacy leakage and potential mitigation strategies~\parencite{wei2020framework}.
They found that batch size, image resolution, choice of activation functions, and the number of local iterations before gradient exchange can impact privacy leakage. 
Supporting findings are reported in \parencite{zhu2019deep, geiping2020inverting, zhao2020idlg, zhu2021r, pan2022exploring}. 
While such parameters and training conditions are certainly relevant and can be carefully selected to prevent attacks, they do not guarantee the protection of sensitive data from \ac{GI} attacks. 
In fact, \citeauthor{geiping2020inverting} demonstrated successful attacks on deep neural networks, \eg ResNet-152, trained for multiple communication rounds and even for batches of $100$ images. 
Moreover, parameter selection is often influenced by other factors, such as hardware limitations. 
Hence, to achieve data privacy, defense mechanisms should be actively developed, analyzed, and applied.

\paragraph{\acf{GP}}
\ac{GP} is the de-facto standard to defend against privacy leakage in collaborative learning scenarios~\parencite{kairouz2021advances}.
The two most common general approaches are gradient compression and noisy gradients.
Gradient compression is primarily used to reduce communication costs and memory usage during collaborative training. 
Since compression mechanisms reduce the entropy of the gradients, they also decrease the success of \ac{GI} attacks. 
This can be achieved with \ac{GP}~\parencite{lin2017deep, tsuzuku2018variance, deng2020model}, by removing the least relevant information from the training gradients before sending them to the server.
Most \sota methods utilize noisy gradients to guarantee a provable degree of privacy for clients in collaborative training~\parencite{bonawitz2017practical, mcmahan2017learning, mcmahan2018general, li2019privacy}.
The concept of using noise to limit the information disclosure about individuals was originally introduced in the field of \ac{DP}~\parencite{abadi2016deep, dwork2014algorithmic}.
In practice, Gaussian or Laplace noise is added to the gradients before their exchange.
Although added noise can suppress \ac{GI} attacks, it also negatively impacts both the training process and the final model utility.
Furthermore, it remains unclear how much privacy can actually be guaranteed in practical scenarios.
Recent work demonstrates that the theoretical privacy guarantees of \ac{DP} do not necessarily ensure practical privacy~\parencite{wang2022differential}.
In general, all gradient perturbation methods exhibit a well-observed side effect: an inherent trade-off between model utility and privacy~\parencite{dwork2014algorithmic, jayaraman2019evaluating, zhu2019deep, wei2020framework, huang2021evaluating}.
While the use of more training data might mitigate such drops in model utility~\parencite{dwork2014algorithmic}, this is often not feasible in practical scenarios.

\paragraph{\acf{PM}}
The authors of~\parencite{scheliga2022precode} propose to extend neural network architectures with \ac{PM}.
Such modules can be generically integrated into any existing model architecture without requiring further model modifications.
More importantly, \acf{PM}s shall not notably harm the final model utility or training process, e.g., by causing increased convergence times.
The authors proposed \ac{PRECODE} as first realization of a \ac{PM}.
\ac{PRECODE} is implemented as \ac{VB} that uses stochastic sampling to generate a stochastic latent representation of the training data, which is then used to calculate the model prediction.
While the model retains feature information that is relevant to solving the models original task, gradients are computed based on stochastic feature representations instead of deterministic ones.
Stochastic feature representations are randomly sampled in each optimization step, which counters iterative \ac{GI} attacks by design.
Recent work has identified severe weaknesses in \ac{PRECODE}'s use of fully connected layers and propose a \acf{CVB} as novel \ac{PM} that uses only convolutional layers for variational modeling~\parencite{scheliga2024privacy}. 
Similar to \ac{PRECODE}, the \ac{CVB} protects against \ac{GI} attacks through stochastic sampling. 
Due to the local connectivity of convolutional layers, the \ac{CVB} also requires considerably less additional parameters. 
This allows for notable reduction of computational and communication costs while effectively preserving privacy. 
An extensive experimental evaluation shows that, compared to other defense mechanisms, \ac{CVB} offers the best trade-off between model utility and privacy in a broad set of scenarios.

\paragraph{Dropout}
A simple and effective method to implement stochastic sampling without additional model parameters is Dropout. 
Dropout is a regularization technique used to reduce overfitting in neural networks~\parencite{hanson1990stochastic, hinton2012improving}. 
While Dropout can enhance the performance of models~\parencite{srivastava2014dropout}, recent studies suggest it can also protect shared gradients from \ac{GI} attacks~\parencite{wei2020framework}. 
However, recent work systematically disseminates Dropout as defense and expose new vulnerabilities~\parencite{scheliga2023dropout}.
They show that an attacker can sufficiently approximate and mimic the stochastic training behavior of the client to effectively bypass the protection seemingly induced by Dropout. 

\section{Preliminaries}
\subsection{Federated Learning}
This paper investigates the privacy of \ac{FL} processes~\parencite{mcmahan2017communication, kairouz2021advances}.
Instead of training one model on a centralized dataset, in \ac{FL} multiple clients collaborate to train a common global model.
Each client has their own private local training dataset.
A centralized orchestration server $S$ coordinates the collaborative training process.
Clients initialize their local model from a given global model state.
After a defined number of local training steps, clients return their local model updates, i.e., their model gradients $G$ to the server.
Hence, training data does not have to be shared and remains local with each client.
The server updates the global model state by aggregating the client's local model updates.
For example, \ac{FedAvg} implements this aggregation using a simple weighted average of the updated states~\parencite{mcmahan2017communication}.
This process is repeated iteratively until convergence or some other termination criterion is satisfied. 
In terms of privacy, all collaborative training approaches that are based on the exchange of gradient information suffer from similar vulnerabilities, e.g., peer-to-peer or cluster-based collaborative training~\parencite{roy2019braintorrent, duan2019astraea, scheliga2024feature}.

\subsection{Threat Model}
\label{sec:threat_model}
Based on the threat model taxonomy in~\parencite{wagner2018technical} and consistent with related work on \ac{GI} attacks~\parencite{zhu2019deep, geiping2020inverting, wei2020framework, scheliga2024privacy, rodriguez2023survey}, we adopt a \emph{honest-but-curious server} threat model.
In this threat model, attackers are \emph{curious} as they apply \ac{GI} attacks to reconstruct potentially sensitive training data from exchanged gradients.
They act \emph{locally} and \emph{internally} on a restricted part of the system, \ie the orchestration server $S$.
Attackers are \emph{passive}, \ie they only read and observe exchanged information, \eg model parameters/gradients.
This passive behavior is also referred to as \emph{honest}, since there is no active interference with the training process.
Consistent with an orchestration server in \ac{FL} scenarios, attackers have \emph{white-box} access to the model and \emph{prior knowledge} on the used training protocols.
Specifically this includes the model $F$, the model weights $W$, the training loss function $\mathcal{L}$, and the exchanged gradients $G = \nabla\mathcal{L}_W(F(x),y)$ of the attacked victim client. 

\subsection{Gradient Inversion Attacks}
\label{sec:prelim_GI}
\acf{GI} attacks aim to reconstruct the training data $(x,y)$ of a victim client $C$ that was used during a local training step of \ac{FL}.
\ac{GI} attacks aim to solve the following optimization problem:
\begin{equation}
    \label{eq:GI}
    \argmin_{ \left( x',y' \right) } \mathrm{GDF}(\nabla \mathcal{L}_W(F(x), y), \nabla \mathcal{L}_W(F(x'),y')) + \vPrior.
\end{equation}
Here $x$ refers to the victim clients' input data and $y$ to the associated ground truth output label.
$F$ is the attacked model with parameters $W$ and $\mathcal{L}_W$ defines the model objective function.
The attack optimization procedure aims to find some dummy data $(x', y')$ that results in similar gradients as the victim clients' data.
The gradient $G=\nabla \mathcal{L}_W(F(x), y)$ denotes the \emph{victim gradient} that is computed using client training data, whereas $G'=\nabla \mathcal{L}_W(F(x'), y')$ is the \emph{dummy gradient} that is computed with dummy data.
$\vPrior$ represents a regularization scheme that aims to improve the reconstruction and stabilize the attack optimization process.
The general working principle of iterative \ac{GI} attacks is formalized in~\autoref{alg:GI}.

\begin{algorithm}[t]
\caption{General \acf{GI} Attack} 
\label{alg:GI}
\textbf{Input}: $F$: neural network; $\mathcal{L}$: training loss function; GDF: gradient distance function;\\
$G=\nabla \mathcal{L}_W(F(x),y)$: victim gradient; $\eta_{\text{GI}}$: learning rate \\
\textbf{Output}: $(x', y')$: training data reconstructions
\begin{algorithmic}[1]
	    \State $x', y' \gets \mathcal{N}(0,\mathcal{I})$ \Comment{{\color{gray} initialize dummy data}}
 		\While{not converged} \Comment{{\color{gray}reiterate until some termination criterion is reached}}
 		    \State $G' \gets \nabla \mathcal{L}_W(F(x'),y')$ \Comment{{\color{gray}calculate dummy gradient}}
 		    \State $\mathrm{GD} \gets \mathrm{GDF}(G, G')$ \Comment{{\color{gray}calculate gradient distance}}
                \State $\mathrm{GD} \gets \mathrm{GD} + \vPrior$ \Comment{{\color{gray}add regularization terms}}
 		    \State $x' \gets x' - \eta_{\text{GI}} \frac{\partial \mathrm{GD}}{\partial x'}$; $y' \gets y' - \eta_{\text{GI}} \frac{\partial \mathrm{GD}}{\partial y'}$ \Comment{{\color{gray}update dummy data}}
 		\EndWhile
 	\State \textbf{return} $(x', y')$
\end{algorithmic}
\end{algorithm}

First the attacker initializes some in- and output dummy data $(x',y')$.
The dummy data is typically initialized randomly from a Normal distribution $\mathcal{N}(0,\mathcal{I})$. 
The attacker iteratively updates the dummy data via gradient descent to minimize a \ac{GDF} between the victim gradient $G$ and the dummy gradient $G'$.
The most commonly used distance functions are the Euclidean and cosine distance.
The attack optimization ends, if a pre-defined termination criterion is reached, \eg a maximum number of iterations or a gradient distance threshold.
Optimization is typically performed using gradient descent with L-BFGS~\parencite{liu1989limited} or Adam~\parencite{kingma2014adam} optimizers. 

The first proposed iterative \ac{GI} attack, \acfi{DLG}, uses Euclidean distance as GDF, no regularization scheme and L-BFGS as optimizer~\parencite{zhu2019deep}.
The attack is mainly limited to the reconstruction of training data from very small batches.
Larger batches make the attack optimization process unstable and reduce the quality of the reconstructions.
\acf{iDLG} extends \ac{DLG} with a deterministic \acf{LRM} to determine the original client labels $y$ if the softmax function is used for activation of the classification layer and cross-entropy loss as a training objective.
Compared to \ac{DLG}, \ac{iDLG} only optimizes for $x'$ and sets $y'=y$ by the \ac{LRM}.
The authors found that this improves convergence speed and reconstruction quality~\parencite{zhao2020idlg}.
The first \ac{GI} attack that reached a significant milestone in terms of reconstruction quality for large input data with more complex neural networks and larger victim batchsizes is \acf{IG}~\parencite{geiping2020inverting}.
Their proposed attack minimizes the gradient distance in terms of layer-wise cosine distance instead of the Euclidean distance.
They use the cosine distance to disentangle the gradient magnitude from its direction.
Intuitively this objective aims to find data $x'$ that results in a similar change in the models prediction, \ie a gradient update $G'$ that follows the same optimization direction as the victim gradient $G$.
Additionally, they apply a regularization scheme based on \acf{TV}~\parencite{rudin1992nonlinear} to increase image fidelity.
The authors furthermore found that using the Adam optimizer~\parencite{kingma2014adam} yields more sophisticated reconstruction results compared to the L-BFGS optimizer.
This is especially the case for deeper models and trained models with an overall smaller gradient magnitude.
With these simple adjustments to the previous attacks, \ac{IG} achieves large improvements in terms of reconstruction quality and is the de-facto standard for empirical evaluation of \ac{GI} attacks and defenses~\parencite{hatamizadeh2022gradvit, gong2023gradient, scheliga2024privacy}.

\subsection{Defense Mechanisms}
\subsubsection{Dropout}
Dropout~\parencite{hanson1990stochastic, hinton2012improving} is a common regularization technique that randomly masks the output of neurons with a chosen probability $\vdrop$.
Each forward pass realizes a different version of the model, making Dropout an effective method for model averaging and in turn prevent models from overfitting~\parencite{srivastava2014dropout}.
Formally, given the output $\vFeaturevector^{(\vlayerindex)}$ of the $\vlayerindex$th layer $\vLayer^{(\vlayerindex)}$ in a model $F$, a subsequent Dropout layer $\vLayer^{(\vlayerindex)}_{\text{dr}}$ multiplies $\vFeaturevector^{(\vlayerindex)}$ element-wise with a random Dropout mask $\psi^{(\vlayerindex)}$ and scales the remaining outputs according to the Dropout rate $\vdrop$ to preserve the output magnitude.
For every Dropout layer $\vLayer^{(\vlayerindex)}_{\text{dr}}$ $\vlayerindex\in \lbrace 1,\dots,\vLayerindex \rbrace$, the Dropout mask $\psi^{(\vlayerindex)}$ is a matrix of independent Bernoulli variables, i.e., $\psi^{(\vlayerindex)} \sim \text{Bernoulli}(\vdrop)$.
We denote $\Psi = \lbrace \psi^{(1)},\dots,\psi^{(\vLayerindex)} \rbrace$ as the set of $\vLayerindex$ random Dropout masks that are applied during a forward pass through the model $F_\Psi$.

\subsubsection{Differential Privacy}
The most common implementation of \ac{DP} for \ac{FL} is \ac{DPSGD}~\parencite{abadi2016deep}.
During local training, the clients calculate the per-sample gradient $g_\vrunningindexi$ for each sample $x_\vrunningindexi$ in an input batch with batchsize $\mathcal{B}$ and scale it by division with 
$\max\left(1, \frac{||g_\vrunningindexi||_2}{\vthreshclip}\right)$.
This scaling is applied to ensure that each single input sample does not have too much influence on the overall gradient.
The clipping threshold $\vthreshclip$ indicates the maximum gradient norm, that should be allowed for each per-sample gradient.
Then, Gaussian noise $\Xi \sim \mathcal{N}(0, \sigma^2 \vthreshclip^2 \mathcal{I})$ is added to the accumulated batch gradient.
To simplify notation, we denote the execution of these steps as  $\Tilde{G}=\text{\ac{DP}}_{\Xi}(G)$.
We use $\Xi_C$ and $\Xi_A$ to refer to the specific noise added by the client and the attacker, respectively.

\subsubsection{Gradient Pruning}
\ac{GP} identifies the gradient elements that carry the least training information, i.e., those with the smallest magnitude, and prunes them to zero.
For a gradient $G=\lbrace G^{(1)},\dots, G^{(\vLayerindex)} \rbrace$, \ac{DGC}~\parencite{lin2017deep} defines the pruning operation as $\Tilde{G}=\text{\ac{GP}}_\Phi(G) = G \odot \Phi$.
Here $\odot$ refers to the layer-wise application of pruning masks $\Phi$.
Elements in the pruning masks have the value 0 for gradient elements with the smallest magnitudes and 1, otherwise.
The pruning masks are determined for each layers gradient based on a pruning ratio $\vprune$.

\subsubsection{Privacy Modules}
Recent work proposed to extend models with \acp{PM} to protect clients from \ac{GI} attacks while maintaining high model utility~\parencite{scheliga2022precode}.
They implement their \ac{PRECODE} module as a \ac{VB} to make feature computation during forwarding stochastic.
This in turn, results in stochastic gradients.
\ac{PRECODE} consists of to fully connected layers: an encoder $E$ and a decoder $D$.
During a forward pass in the model, the encoder $E$ encodes a latent feature representation $\vFeaturevector$ as $E(\vFeaturevector) = \lbrack \mu_E, \sigma_E \rbrack$.
Here $\vFeaturevector=I(x)$ represents the latent representations computed by forward propagating an input sample $x$ through all layers of the base model prior to the \ac{PRECODE} module.
Next, a compressed bottleneck representation $\vVBbottlerep \sim \mathcal{N}(\mu_E, \sigma_E)$ is randomly sampled.
Since direct sampling from $\mathcal{N}(\mu_E, \sigma_E)$ is a non-differentiable operation with respect to $\mu_E$ and $\sigma_E$ they apply the reparameterization trick described in~\parencite{kingma2013auto}: $\vVBbottlerep = \mu_E + \sigma_E \odot \vVBeps$.
Here $\vVBeps\sim\mathcal{N}(0, \mathcal{I})$ is randomly sampled from a standard normal distribution and $\odot$ denotes element-wise multiplication.
Finally, the decoder generates a new stochastic representation $D(\vVBbottlerep)=\hat{\vFeaturevector}$ which is then used to calculate the model prediction $\hat{y} = O(\hat{\vFeaturevector})$.
$O$ refers to the layer(s) of the model that follow the \ac{VB}.
The \ac{CVB} proposed in~\parencite{scheliga2024privacy} functions in the same way, but only uses convolutional layers instead of fully connected layers.
Since convolutional layers make use of parameter-sharing, the number of additional model parameters can be reduced and privacy preserving effects improved.

\section{Targeted Attacks -- Mimicking Client Behavior}
Recent work investigated the privacy preserving effects of Dropout and found that stochastic sampling of the Dropout masks disables the reconstruction of training data through general \ac{GI} attacks such as \ac{IG}~\parencite{scheliga2023dropout}.
Without Dropout, a model $F$ produces deterministic outputs for given inputs.
Dropout introduces randomness, making the model $F_\Psi$ stochastic. 
In each training step new Dropout masks are sampled.
This results in different versions of the model, which leads to stochastic predictions and gradients during training.
Since an attacker does not have knowledge of the specific Dropout masks $\Psi_C$ that a client $C$ used during training, the attacker applies randomly sampled Dropout masks $\Psi_A$ for the attack.
Consequently, the dummy gradients $G' = \nabla \mathcal{L}_W(F_{\Psi_A}(x'),y'))$ vary significantly in each attack iteration and will not match the client gradient $G = \nabla \mathcal{L}_W(F_{\Psi_C}(x), y))$.
As a result, the attacker is unable to compute sufficient reconstructions~\parencite{scheliga2023dropout}.
In response they propose a \acf{DIA} that approximates $F_{\Psi_A} \approx F_{\Psi_C}$ by joint optimization of $\Psi_A$ during the \ac{GI} attack.
If a sufficient approximation is found, the attack can mimic the gradient computation process of the client and therefore bypass the protection of Dropout.
In the following section we adopt this attack principle and show how it can be applied to other defense mechanisms.
In particular, we will empirically demonstrate, that if an attacker can sufficiently approximate and mimic the gradient computation process of the client, the protection of \ac{DP}, \ac{GP}, \ac{PRECODE} and \ac{CVB} can be bypassed.

Throughout the following sections we show experimental results for some preliminary experiments to provide an empirical proof of concept for the performance of the described attacks.
We compare the \ac{IG} attack with our proposed targeted attack variations for a \ac{CNN} and \ac{ViT} on the MNIST~\parencite{lecun1998gradient} and CIFAR-10~\parencite{krizhevsky2009learning} datasets.
We measure reconstruction quality in terms of \ac{SSIM} and \acf{ASR}.
Details on the experimental setup for all reconstruction experiments in this paper are presented in~\autoref{sec:design}.

\subsection{Differential Privacy Inversion Attack}
Unlike Dropout, \ac{DP} does not create different model realizations, as both the client and attacker use the same deterministic model $F$ for gradient computation. 
Instead, \ac{DP} adds randomness directly to the training gradients, causing different realizations of the exchanged gradients. 
An attacker aims to find dummy images $x'$ so that the dummy gradients closely match the client's original training gradients, i.e., $G' = G$. 
However, the client sends the perturbed gradients $\Tilde{G}$ to the server. 
Therefore, the attacker must optimize the distance between the dummy gradients $G'$ and the perturbed client gradients $\Tilde{G}$. 
Since $\Tilde{G}=\text{\ac{DP}}_{\Xi_C}(G)$ is inherently different compared to $G$, reconstruction quality decreases with larger differences between $\Tilde{G}$ and $G$.

While \ac{DP} can protect from \ac{GI} attacks when sufficient noise is added, previous studies show that low levels of noise are insufficient to prevent data reconstruction~\parencite{huang2021evaluating, scheliga2024privacy}. 
However, we argue, that even with high noise levels, if the attacker has access to the specific noise $\Xi_C$ added by the client, they could still reconstruct the training data. 
This is because the attacker could fully mimic the gradient computation process, using $\text{\ac{GDF}}(\Tilde{G}, \Tilde{G'})$ for optimization, where $\Tilde{G}=\text{\ac{DP}}_{\Xi_C}(G)$ and $\Tilde{G'}=\text{\ac{DP}}_{\Xi_C}(G')$. 
We test this assumption in by giving the attacker access to the client's noise, i.e., $\text{\ac{DP}}_{\Xi_A} = \text{\ac{DP}}_{\Xi_C}$.
Consistent with the \emph{well-informed attack} on Dropout in~\parencite{scheliga2023dropout}, we refer to this attack as \ac{WIIG}.
Based on a set of preliminary experiments, we apply \ac{DP} with parameters $\vthreshclip = 20$ and $\sigma = 10^{-3}$.
This level of noise noticeably impacted the reconstruction quality of the \ac{IG} attack while minimizing the impact on model utility.

\begin{table*}[h!]
\centering
\caption{
Privacy metrics for \ac{IG}, \ac{WIIG} and \ac{DPIA} for training gradients defended with \ac{DP}.
The training gradients are attacked for the \ac{CNN} and \ac{ViT} on the MNIST and CIFAR-10 datasets.
Arrows indicate direction of improvement from the viewpoint of a defending client.
Bold and italic formatting highlight best and worst results, respectively.
}
\label{tab:dpia_results_sdia}
\begin{tabular}{c|c|c|c|c|c}
\toprule
 & Model & \ac{DP} $(\vthreshclip, \sigma)$ & Attack & \ac{SSIM} $\downarrow$ & \ac{ASR} [\%] $\downarrow$ \\
\midrule
\multirow[c]{8}{*}{\rotatebox{90}{MNIST}} & \multirow[c]{4}{*}{\ac{CNN}} & - & \ac{IG} & {\cellcolor[HTML]{BD1726}} \color[HTML]{F1F1F1} \itshape 0.95 ($\pm$0.06) & {\cellcolor[HTML]{A50026}} \color[HTML]{F1F1F1} \itshape 100 \\
\cline{3-6}
 &  & \multirow[c]{3}{*}{$(20, 10^{-3})$} & \ac{IG} & {\cellcolor[HTML]{FECC7B}} \color[HTML]{000000} 0.64 ($\pm$0.11) & {\cellcolor[HTML]{E44C34}} \color[HTML]{F1F1F1} 85.16 \\
 &  &  & \ac{WIIG} & {\cellcolor[HTML]{D62F27}} \color[HTML]{F1F1F1}  0.90 ($\pm$0.06) & {\cellcolor[HTML]{A50026}} \color[HTML]{F1F1F1} \itshape 100 \\
 &  &  & \ac{DPIA} & {\cellcolor[HTML]{A0D669}} \color[HTML]{000000} \bfseries 0.29 ($\pm$0.13) & {\cellcolor[HTML]{0E8245}} \color[HTML]{F1F1F1} \bfseries 5.47 \\
\cline{2-6}
 & \multirow[c]{4}{*}{\ac{ViT}} & - & \ac{IG}  & {\cellcolor[HTML]{A90426}} \color[HTML]{F1F1F1} \itshape 0.99 ($\pm$0.00) & {\cellcolor[HTML]{A50026}} \color[HTML]{F1F1F1} \itshape 100 \\
\cline{3-6}
& & \multirow[c]{3}{*}{$(20, 10^{-3})$} & \ac{IG} & {\cellcolor[HTML]{FFF0A6}} \color[HTML]{000000} 0.55 ($\pm$0.09) & {\cellcolor[HTML]{F7844E}} \color[HTML]{F1F1F1} 76.56 \\
 &  &  & \ac{WIIG} & {\cellcolor[HTML]{C21C27}} \color[HTML]{F1F1F1}  0.94 ($\pm$0.03) & {\cellcolor[HTML]{A50026}} \color[HTML]{F1F1F1} \itshape 100 \\
 &  &  & \ac{DPIA} & {\cellcolor[HTML]{148E4B}} \color[HTML]{F1F1F1} \bfseries 0.08 ($\pm$0.06) & {\cellcolor[HTML]{006837}} \color[HTML]{F1F1F1} \bfseries 0 \\

 \midrule
 
 \multirow[c]{8}{*}{\rotatebox{90}{CIFAR-10}} & \multirow[c]{4}{*}{\ac{CNN}} & - & \ac{IG} & {\cellcolor[HTML]{E0422F}} \color[HTML]{F1F1F1} \itshape 0.87 ($\pm$0.11) & {\cellcolor[HTML]{B30D26}} \color[HTML]{F1F1F1} 96.88 \\
\cline{3-6}
& & \multirow[c]{3}{*}{$(20, 10^{-3})$} & \ac{IG} & {\cellcolor[HTML]{ABDB6D}} \color[HTML]{000000} 0.31 ($\pm$0.11) & {\cellcolor[HTML]{0E8245}} \color[HTML]{F1F1F1} 5.47 \\
 &  &  & \ac{WIIG} & {\cellcolor[HTML]{EE613E}} \color[HTML]{F1F1F1}  0.82 ($\pm$0.06) & {\cellcolor[HTML]{A50026}} \color[HTML]{F1F1F1} \itshape 100 \\
 &  &  & \ac{DPIA} & {\cellcolor[HTML]{8CCD67}} \color[HTML]{000000} \bfseries 0.26 ($\pm$0.10) & {\cellcolor[HTML]{036E3A}} \color[HTML]{F1F1F1} \bfseries 1.56 \\
\cline{2-6}
 & \multirow[c]{4}{*}{\ac{ViT}} & - & \ac{IG} & {\cellcolor[HTML]{D62F27}} \color[HTML]{F1F1F1} \itshape 0.90 ($\pm$0.05) & {\cellcolor[HTML]{A50026}} \color[HTML]{F1F1F1} \itshape 100 \\
\cline{3-6}
& & \multirow[c]{3}{*}{$(20, 10^{-3})$} & \ac{IG} & {\cellcolor[HTML]{6BBF64}} \color[HTML]{000000} 0.21 ($\pm$0.07) & {\cellcolor[HTML]{006837}} \color[HTML]{F1F1F1} \bfseries 0 \\
 &  &  & \ac{WIIG} & {\cellcolor[HTML]{F67F4B}} \color[HTML]{F1F1F1} 0.77 ($\pm$0.08) & {\cellcolor[HTML]{A50026}} \color[HTML]{F1F1F1} \itshape 100 \\
 &  &  & \ac{DPIA} & {\cellcolor[HTML]{17934E}} \color[HTML]{F1F1F1} \bfseries 0.09 ($\pm$0.04) & {\cellcolor[HTML]{006837}} \color[HTML]{F1F1F1} \bfseries 0 \\
\bottomrule
\end{tabular}
\end{table*}

\autoref{tab:dpia_results_sdia} shows the results of these experiments.
Compared to an unprotected baseline model, \ac{DP} reduces the reconstruction quality under the \ac{IG} attack. 
For the \ac{CNN}, \ac{IG} achieves an \ac{ASR} of $85.16\%$ on MNIST and $5.47\%$ on CIFAR-10. 
For the \ac{ViT}, the \ac{ASR} is $76.56\%$ and $0\%$, respectively. 
As expected, \ac{WIIG} increases the \ac{ASR} to $100\%$ across all models and datasets, as it replicates the exact gradient computation process performed by the client during local training. 
Therefore, if an attacker can sufficiently approximate the noise and scaling factor, reconstruction quality should improve compared to a baseline \ac{IG} attack.

In reality, the attacker does not have access to $\text{\ac{DP}}_{\Xi_C}$. 
However, similar to the Dropout mask approximations in \ac{DIA}~\parencite{scheliga2023dropout}, the attacker can attempt to approximate the client's noise, aiming for $\text{\ac{DP}}_{\Xi_A} \approx \text{\ac{DP}}_{\Xi_C}$, i.e., $\Xi_A \approx \Xi_C$. 
We implement this approach as a \ac{DPIA}. 
In addition to the general \ac{GI} attack, the attacker randomly initializes dummy noise $\Xi_A = \lbrace \xi_A^{(\vlayerindex)} \sim \mathcal{N}(0, \sigma^2 \vthreshclip^2 \mathcal{I}) | \forall \vlayerindex=1,\dots,\vLayerindex \rbrace$. 
During the attack, this dummy noise is applied to the dummy gradients, i.e., $\Tilde{G'} = \text{\ac{DP}}_{\Xi_A}(G')$, and the gradient distance is calculated between the perturbed client gradient $\Tilde{G}$ and the dummy perturbed gradients $\Tilde{G'}$. 
In each attack iteration, the attacker adjusts the dummy noise based on the gradient distance with respect to the dummy noise. 
Additionally, we regularize the dummy noise to have a mean of 0 and a standard deviation of $\sigma^2 \vthreshclip^2$.
If the attacker can sufficiently approximate the noise such that $\Xi_A\approx\Xi_C$, we expect the reconstruction quality of the dummy data to improve. 

However, the results in \autoref{tab:dpia_results_sdia} indicate that \ac{DPIA} does not find sufficient approximations $\Xi_A$.
In fact, the reconstruction quality using \ac{DPIA} is significantly lower than with \ac{IG}, as indicated by the reduced \ac{SSIM} values and an \ac{ASR} close to or equal to $0\%$. 
While \ac{DPIA} computes dummy gradients $\Tilde{G'}$ that closely match the victim’s perturbed gradients $\Tilde{G}$, this does not translate to high-quality reconstructions. 
Since \ac{DPIA} jointly optimizes the dummy noise and the dummy data by minimizing the \ac{GDF}, there are too many possibilities for the dummy noise to compensate for misalignments between the victim’s and dummy’s gradients. 
In the worst case, \ac{DPIA} may only be optimizing for an optimal dummy noise to minimize the gradient distance, while the dummy data remains of poor quality.
Although we were unable to sufficiently approximate the noise applied by the client using \ac{DP}, and thus did not improve reconstruction quality, it is important to note that \ac{DP} always involves a trade-off between model utility and privacy.
While \ac{WIIG} achieves an \ac{ASR} of $100\%$, this was only considered as a proof of concept. 
In a realistic scenario, an attacker would not have access to the ground truth noise applied by the client.

\subsection{Gradient Pruning Inversion Attack}
Unlike Dropout, \ac{DP}, \ac{PRECODE}, and \ac{CVB}, \ac{GP} does not affect gradient computation through stochastic processes. 
Instead, it deterministically prunes gradient values based on their magnitude. 
Intuitively the main defensive capability of \ac{GP} against \ac{GI} attacks is the reduced amount of gradient information available for attack optimization. 
However, since only the lowest magnitude values are typically pruned, information on the most important feature activations in the model is still retained.
This suggests that even with high pruning rates, this information could be sufficient for \ac{GI} attacks.
We argue that the primary reason for the decrease in reconstruction quality when \ac{GP} is applied is that the attacker does not mimic the pruning step used by the client during gradient computation. 
Without applying a similar pruning regimen, it becomes challenging, if not impossible, for the attacker to find dummy data $x'$ that results in a dummy gradient $G'$ that is as sparse as the pruned client gradient $\Tilde{G} = \text{\ac{GP}}_{\Phi_C}(G)$.
Fortunately for the attacker, the client reveals the pruning masks used during local training. 
The ground truth pruning masks $\Phi_C = \lbrace \phi^{(1)}_C, \dots, \phi^{(\vLayerindex)}_C \rbrace$ can be determined as follows:
\begin{equation}
    \label{eq:prune_mask_reconstruction}
    \phi^{(\vlayerindex)}_C = \mathbf{1}\left( \text{abs} \left(G^{(\vlayerindex)}\right) > 0\right) \forall \vlayerindex=1,\dots,\vLayerindex.
\end{equation}
Here, $\mathbf{1}$ denotes that elements in $\phi^{(\vlayerindex)}_C$ are set to 1 if the corresponding gradient elements meet the condition and are set to 0 otherwise.
We implement a \ac{GPIA} that takes advantage of this knowledge by applying the client’s pruning mask $\Phi_C$ to the dummy gradients $G'$ during the attack optimization, i.e., $\Tilde{G'} = \text{\ac{GP}}_{\Phi_C}(G')$. 
As a result, the zero values in $\Tilde{G}$ and $\Tilde{G'}$ match by default.
Previous studies indicate that high pruning rates $\vprune$ are required to have any significant impact on privacy preservation~\parencite{scheliga2024privacy}. 
This means that typically at least $90\%$ of the values in $\Tilde{G}$ and $\Tilde{G'}$ match by default. 
Since $\Tilde{G}$ retains the most informative gradient values, i.e., those with the highest magnitude, these might be sufficient to reconstruct the client’s training data $x$ from the dummy data $x'$.

\begin{table*}[t!]
\centering
\caption{
Privacy metrics for \ac{IG} and \ac{GPIA} for training gradients defended with \ac{GP} with increasing pruning rates $\vprune$.
The training gradients are attacked for the \ac{CNN} and \ac{ViT} on the MNIST and CIFAR-10 datasets.
Arrows indicate direction of improvement from the viewpoint of a defending client.
Bold and italic formatting highlight best and worst results, respectively.
}
\label{tab:gpia_results_sdia}
\begin{tabular}{c|c|c|c|c|c}
\toprule
 & Model & \ac{GP} $\vprune$ & Attack & \ac{SSIM} $\downarrow$ & \ac{ASR} [\%] $\downarrow$ \\
\midrule
\multirow[c]{14}{*}{\rotatebox{90}{MNIST}} & \multirow[c]{7}{*}{\ac{CNN}} & - & \ac{IG} & {\cellcolor[HTML]{BD1726}} \color[HTML]{F1F1F1} \itshape 0.95 ($\pm$0.06) & {\cellcolor[HTML]{A50026}} \color[HTML]{F1F1F1} \itshape 100 \\
\cline{3-6}
&  & \multirow[c]{2}{*}{0.90} & \ac{IG} & {\cellcolor[HTML]{FED07E}} \color[HTML]{000000} 0.63 ($\pm$0.09) & {\cellcolor[HTML]{D83128}} \color[HTML]{F1F1F1} 89.84 \\
 &  &  & \ac{GPIA} & {\cellcolor[HTML]{D22B27}} \color[HTML]{F1F1F1} 0.91 ($\pm$0.07) & {\cellcolor[HTML]{A50026}} \color[HTML]{F1F1F1} \itshape 100 \\
 \cline{3-6}
 &  & \multirow[c]{2}{*}{0.95} & \ac{IG} & {\cellcolor[HTML]{FFF8B4}} \color[HTML]{000000} 0.52 ($\pm$0.08) & {\cellcolor[HTML]{FECC7B}} \color[HTML]{000000} 64.06 \\
 &  &  & \ac{GPIA} & {\cellcolor[HTML]{EE613E}} \color[HTML]{F1F1F1} 0.82 ($\pm$0.12) & {\cellcolor[HTML]{A70226}} \color[HTML]{F1F1F1} 99.22 \\
 \cline{3-6}
 &  & \multirow[c]{2}{*}{0.99} & \ac{IG} & {\cellcolor[HTML]{7FC866}} \color[HTML]{000000} \bfseries 0.24 ($\pm$0.05) & {\cellcolor[HTML]{006837}} \color[HTML]{F1F1F1} \bfseries 0 \\
 &  &  & \ac{GPIA} & {\cellcolor[HTML]{FFFEBE}} \color[HTML]{000000} 0.50 ($\pm$0.13) & {\cellcolor[HTML]{FDFEBC}} \color[HTML]{000000} 49.22 \\
 \cline{2-6}
 & \multirow[c]{7}{*}{\ac{ViT}} & - & \ac{IG}  & {\cellcolor[HTML]{A90426}} \color[HTML]{F1F1F1} \itshape 0.99 ($\pm$0.00) & {\cellcolor[HTML]{A50026}} \color[HTML]{F1F1F1} \itshape 100 \\
\cline{3-6}
& & \multirow[c]{2}{*}{0.90} & \ac{IG} & {\cellcolor[HTML]{66BD63}} \color[HTML]{F1F1F1} 0.20 ($\pm$0.06) & {\cellcolor[HTML]{006837}} \color[HTML]{F1F1F1} \bfseries 0 \\
 &  &  & \ac{GPIA} & {\cellcolor[HTML]{B91326}} \color[HTML]{F1F1F1} 0.96 ($\pm$0.02) & {\cellcolor[HTML]{A50026}} \color[HTML]{F1F1F1} \itshape 100 \\
 \cline{3-6}
 &  & \multirow[c]{2}{*}{0.95} & \ac{IG} & {\cellcolor[HTML]{07753E}} \color[HTML]{F1F1F1} 0.03 ($\pm$0.02) & {\cellcolor[HTML]{006837}} \color[HTML]{F1F1F1} \bfseries 0 \\
 &  &  & \ac{GPIA} & {\cellcolor[HTML]{C62027}} \color[HTML]{F1F1F1} 0.93 ($\pm$0.03) & {\cellcolor[HTML]{A50026}} \color[HTML]{F1F1F1} \itshape 100 \\
 \cline{3-6}
 &  & \multirow[c]{2}{*}{0.99} & \ac{IG} & {\cellcolor[HTML]{026C39}} \color[HTML]{F1F1F1} \bfseries 0.01 ($\pm$0.02) & {\cellcolor[HTML]{006837}} \color[HTML]{F1F1F1} \bfseries 0 \\
 &  &  & \ac{GPIA} & {\cellcolor[HTML]{E24731}} \color[HTML]{F1F1F1} 0.86 ($\pm$0.05) & {\cellcolor[HTML]{A50026}} \color[HTML]{F1F1F1} \itshape 100 \\

 \midrule
 
 \multirow[c]{14}{*}{\rotatebox{90}{CIFAR-10}} & \multirow[c]{7}{*}{\ac{CNN}} & - & \ac{IG} & {\cellcolor[HTML]{E0422F}} \color[HTML]{F1F1F1} \itshape 0.87 ($\pm$0.11) & {\cellcolor[HTML]{B30D26}} \color[HTML]{F1F1F1} \itshape 96.88 \\
\cline{3-6}
& & \multirow[c]{2}{*}{0.90} & \ac{IG} & {\cellcolor[HTML]{F7FCB4}} \color[HTML]{000000} 0.48 ($\pm$0.08) & {\cellcolor[HTML]{F4FAB0}} \color[HTML]{000000} 46.88 \\
 &  &  & \ac{GPIA} & {\cellcolor[HTML]{FDAD60}} \color[HTML]{000000} 0.70 ($\pm$0.12) & {\cellcolor[HTML]{D42D27}} \color[HTML]{F1F1F1} 90.62 \\
 \cline{3-6}
 &  & \multirow[c]{2}{*}{0.95} & \ac{IG} & {\cellcolor[HTML]{AFDD70}} \color[HTML]{000000} 0.32 ($\pm$0.07) & {\cellcolor[HTML]{006837}} \color[HTML]{F1F1F1} \bfseries 0 \\
 &  &  & \ac{GPIA} & {\cellcolor[HTML]{FFF6B0}} \color[HTML]{000000} 0.53 ($\pm$0.15) & {\cellcolor[HTML]{FDBF6F}} \color[HTML]{000000} 66.41 \\
 \cline{3-6}
 &  & \multirow[c]{2}{*}{0.99} & \ac{IG} & {\cellcolor[HTML]{3FAA59}} \color[HTML]{F1F1F1} \bfseries 0.15 ($\pm$0.04) & {\cellcolor[HTML]{006837}} \color[HTML]{F1F1F1} \bfseries 0 \\
 &  &  & \ac{GPIA} & {\cellcolor[HTML]{A0D669}} \color[HTML]{000000} 0.29 ($\pm$0.13) & {\cellcolor[HTML]{0A7B41}} \color[HTML]{F1F1F1} 3.91 \\
 \cline{2-6}
 & \multirow[c]{7}{*}{\ac{ViT}} & - & \ac{IG} & {\cellcolor[HTML]{D62F27}} \color[HTML]{F1F1F1} \itshape 0.90 ($\pm$0.05) & {\cellcolor[HTML]{A50026}} \color[HTML]{F1F1F1} \itshape 100 \\
\cline{3-6}
& & \multirow[c]{2}{*}{0.90} & \ac{IG} & {\cellcolor[HTML]{17934E}} \color[HTML]{F1F1F1} 0.09 ($\pm$0.04) & {\cellcolor[HTML]{006837}} \color[HTML]{F1F1F1} \bfseries 0 \\
 &  &  & \ac{GPIA} & {\cellcolor[HTML]{F67F4B}} \color[HTML]{F1F1F1} 0.77 ($\pm$0.07) & {\cellcolor[HTML]{A50026}} \color[HTML]{F1F1F1} \itshape 100 \\
 \cline{3-6}
 &  & \multirow[c]{2}{*}{0.95} & \ac{IG} & {\cellcolor[HTML]{05713C}} \color[HTML]{F1F1F1} 0.02 ($\pm$0.02) & {\cellcolor[HTML]{006837}} \color[HTML]{F1F1F1} \bfseries 0 \\
 &  &  & \ac{GPIA} & {\cellcolor[HTML]{FCA85E}} \color[HTML]{000000} 0.71 ($\pm$0.08) & {\cellcolor[HTML]{A50026}} \color[HTML]{F1F1F1} \itshape 100 \\
 \cline{3-6}
 &  & \multirow[c]{2}{*}{0.99} & \ac{IG} & {\cellcolor[HTML]{026C39}} \color[HTML]{F1F1F1} \bfseries 0.01 ($\pm$0.01) & {\cellcolor[HTML]{006837}} \color[HTML]{F1F1F1} \bfseries 0 \\
 &  &  & \ac{GPIA} & {\cellcolor[HTML]{FEE28F}} \color[HTML]{000000} 0.59 ($\pm$0.09) & {\cellcolor[HTML]{EA5739}} \color[HTML]{F1F1F1} 83.59 \\
\bottomrule
\end{tabular}
\end{table*}

We evaluate the performance of \ac{GPIA} on \ac{GP}-protected gradients and compare it to the \ac{IG} attack. 
We do not explicitly consider a \ac{WIIG} attacker, as \ac{GPIA} computes the ground truth pruning masks, which are inherently part of the exchanged training gradients.
We test both attacks for increasing pruning ratios $\vprune \in \lbrace 0.90, 0.95, 0.99\rbrace$ to determine the level of gradient sparsity required to effectively prevent the reconstruction of training data.
The results of these experiments are shown in \autoref{tab:gpia_results_sdia}.
For both datasets, models, and all considered pruning rates $\vprune$, \ac{GPIA} consistently improves reconstruction quality compared to \ac{IG}, as indicated by higher \ac{SSIM} and \ac{ASR}. 
As expected, since only the gradients with the smallest magnitudes are pruned, enough information for reconstruction often remains. 
By mimicking the full gradient computation process, including the application of \ac{GP}, rather than just gradient computation as in \ac{IG}, the \ac{GPIA} attacker can better optimize the dummy data to match the victim’s gradients.
However, as the pruning rate increases, \ac{ASR} decreases for the \ac{CNN}. 
For a pruning rate of $\vprune=0.95$, \ac{GPIA} achieves an \ac{ASR} of $99.22\%$ and $66.41\%$ for MNIST and CIFAR-10, respectively, while a pruning rate of $\vprune=0.99$ results in \ac{ASR} values of $49.22\%$ and $3.91\%$ for the same datasets.
For the \ac{ViT}, this decrease in reconstruction quality requires higher pruning rates and more complex data. 
For MNIST, \ac{GPIA} achieves an \ac{ASR} of $100\%$ for all pruning rates, compared to $0\%$ under \ac{IG}. 
For CIFAR-10, the \ac{ASR} can be reduced to $83.59\%$ with a pruning rate of $\vprune=0.99$. 
We suspect that for the \ac{ViT}, reconstruction quality is less affected because the overall number of parameters is significantly larger compared to our considered \ac{CNN}. 
For example, with a large pruning rate of $\vprune=0.99$, $660$ and $16,092$ non-zero gradient values remain for attack optimization for the \ac{CNN} and \ac{ViT}, respectively. 
Depending on the size of the model, even at high pruning rates there is still enough gradient information left to achieve good reconstructions.

\subsection{Variational Bottleneck Inversion Attack}
Similar to Dropout the defensive capabilities of the \ac{VB}-based \acp{PM} \ac{PRECODE} and \ac{CVB} come from the randomness induced by the stochastic sampling during the forwarding step.
They turn the deterministic model $F$ into a stochastic one, that depends on the sampled values of $\vVBeps$, which are required for the reparameterization trick during model training.
Since $\vVBeps$ is randomly sampled in each optimization step, small changes in the dummy image $x'$ lead to an increased entropy of the dummy's latent representations $D(\vVBbottlerep)=\hat{\vFeaturevector}$.
This makes it difficult for the optimizer to find dummy images $x'$ that minimize the reconstruction loss.

Hence, to sufficiently mimic the client gradient computation process, an attacker has to approximate the randomly sampled reparameterization matrix $\varepsilon_{\text{VB}_C}$.
We implement \ac{VBIA} accordingly, by jointly optimizing for $\varepsilon_{\text{VB}_C}$ during the attack optimization process.
In addition to the general \ac{GI} attack, the attacker randomly initializes a dummy reparameterization matrix $\varepsilon_{\text{VB}_A} \sim \mathcal{N}(0, \mathcal{I})$. 
During forward passes in the attack optimization process, this dummy reparameterization matrix is used instead of a randomly sampled one in the \ac{VB}. 
In each attack iteration, the attacker updates both the dummy data and the dummy reparameterization matrix $\varepsilon_{\text{VB}_A}$ based on the gradient distance GD. 
If the attacker can sufficiently approximate the reparameterization matrix such that $\varepsilon_{\text{VB}_A} \approx \varepsilon_{\text{VB}_C}$, the reconstruction quality of the dummy data is expected to improve.

We evaluate the performance of \ac{VBIA} on \ac{PRECODE} and \ac{CVB} protected gradients and compare it to the \acf{PIG} ("Ignore attack") proposed in~\parencite{scheliga2024privacy}.
By excluding the most stochastic gradients from gradient distance calculation during attack optimization, \ac{PIG} was shown to achieve better reconstructions for \ac{PRECODE} and \ac{CVB} compared to \ac{IG}. 
For comparison, we evaluate a \ac{WIIG} attack, where the attacker has access to the true reparameterization matrix $\varepsilon_{\text{VB}_C}$ sampled by the client during training, i.e., $\varepsilon_{\text{VB}_A} = \varepsilon_{\text{VB}_C}$. 
In reality, however, the attacker does not have this knowledge and must approximate $\varepsilon_{\text{VB}_A} \approx \varepsilon_{\text{VB}_C}$.
We parameterize \ac{PRECODE} and \ac{CVB} as recommended in~\parencite{scheliga2024privacy}.

\begin{table*}[b!]
\centering
\caption{
Privacy metrics for \ac{PIG}, \ac{WIIG} and \ac{VBIA}.
The models use either \ac{PRECODE} or a \ac{CVB} as privacy module.
The training gradients are attacked for the \ac{CNN} and \ac{ViT} on the MNIST and CIFAR-10 datasets.
Arrows indicate direction of improvement from the viewpoint of a defending client.
Bold and italic formatting highlight best and worst results, respectively.
}
\label{tab:vbia_results_sdia}
\begin{tabular}{c|c|c|c|c|c}
\toprule
 & Model & Defense & Attack & \ac{SSIM} $\downarrow$ & \ac{ASR} [\%] $\downarrow$ \\
\midrule
\multirow[c]{14}{*}{\rotatebox{90}{MNIST}} & \multirow[c]{7}{*}{\ac{CNN}} & - & \ac{IG} & {\cellcolor[HTML]{BD1726}} \color[HTML]{F1F1F1} \itshape 0.95 ($\pm$0.06) & {\cellcolor[HTML]{A50026}} \color[HTML]{F1F1F1} \itshape 100 \\
\cline{3-6}
&  & \multirow[c]{3}{*}{\ac{PRECODE}} & \ac{PIG} & {\cellcolor[HTML]{E0F295}} \color[HTML]{000000} 0.42 ($\pm$0.09) & {\cellcolor[HTML]{6BBF64}} \color[HTML]{000000} 21.09 \\
&  &  & \ac{WIIG} & {\cellcolor[HTML]{E0422F}} \color[HTML]{F1F1F1} 0.87 ($\pm$0.13) & {\cellcolor[HTML]{AB0626}} \color[HTML]{F1F1F1} 98.44 \\
 &  &  & \ac{VBIA} & {\cellcolor[HTML]{BD1726}} \color[HTML]{F1F1F1} \itshape 0.95 ($\pm$0.07) & {\cellcolor[HTML]{A50026}} \color[HTML]{F1F1F1} \itshape 100 \\
\cline{3-6}
&  & \multirow[c]{3}{*}{\ac{CVB}} & \ac{PIG} & {\cellcolor[HTML]{A0D669}} \color[HTML]{000000} \bfseries 0.29 ($\pm$0.09) & {\cellcolor[HTML]{016A38}} \color[HTML]{F1F1F1} \bfseries 0.78 \\
&  &  & \ac{WIIG} & {\cellcolor[HTML]{BD1726}} \color[HTML]{F1F1F1} \itshape 0.95 ($\pm$0.04) & {\cellcolor[HTML]{A50026}} \color[HTML]{F1F1F1} \itshape 100 \\
 &  &  & \ac{VBIA} & {\cellcolor[HTML]{DC3B2C}} \color[HTML]{F1F1F1} 0.88 ($\pm$0.09) & {\cellcolor[HTML]{A70226}} \color[HTML]{F1F1F1} 99.22 \\
\cline{2-6}
& \multirow[c]{7}{*}{\ac{ViT}} & - & \ac{IG}  & {\cellcolor[HTML]{A90426}} \color[HTML]{F1F1F1} 0.99 ($\pm$0.00) & {\cellcolor[HTML]{A50026}} \color[HTML]{F1F1F1} \itshape 100 \\
\cline{3-6}
& & \multirow[c]{3}{*}{\ac{PRECODE}} & \ac{PIG} & {\cellcolor[HTML]{0A7B41}} \color[HTML]{F1F1F1} \bfseries 0.04 ($\pm$0.02) & {\cellcolor[HTML]{006837}} \color[HTML]{F1F1F1} \bfseries 0 \\
&  &  & \ac{WIIG} & {\cellcolor[HTML]{A90426}} \color[HTML]{F1F1F1} 0.99 ($\pm$0.00) & {\cellcolor[HTML]{A50026}} \color[HTML]{F1F1F1} \itshape 100 \\
 &  &  & \ac{VBIA} & {\cellcolor[HTML]{C21C27}} \color[HTML]{F1F1F1} 0.94 ($\pm$0.03) & {\cellcolor[HTML]{A50026}} \color[HTML]{F1F1F1} \itshape 100 \\
\cline{3-6}
&  & \multirow[c]{3}{*}{\ac{CVB}} & \ac{PIG} & {\cellcolor[HTML]{C9E881}} \color[HTML]{000000} 0.37 ($\pm$0.08) & {\cellcolor[HTML]{118848}} \color[HTML]{F1F1F1} 7.03 \\
&  &  & \ac{WIIG} & {\cellcolor[HTML]{A50026}} \color[HTML]{F1F1F1} \itshape 1.00 ($\pm$0.00) & {\cellcolor[HTML]{A50026}} \color[HTML]{F1F1F1} \itshape 100 \\
 &  &  & \ac{VBIA} & {\cellcolor[HTML]{A50026}} \color[HTML]{F1F1F1} \itshape 1.00 ($\pm$0.00) & {\cellcolor[HTML]{A50026}} \color[HTML]{F1F1F1} \itshape 100 \\

\midrule
 
\multirow[c]{14}{*}{\rotatebox{90}{CIFAR-10}} & \multirow[c]{7}{*}{\ac{CNN}} & - & \ac{IG} & {\cellcolor[HTML]{E0422F}} \color[HTML]{F1F1F1} 0.87 ($\pm$0.11) & {\cellcolor[HTML]{B30D26}} \color[HTML]{F1F1F1} 96.88 \\
\cline{3-6}
& & \multirow[c]{3}{*}{\ac{PRECODE}} & \ac{PIG} & {\cellcolor[HTML]{AFDD70}} \color[HTML]{000000} 0.32 ($\pm$0.09) & {\cellcolor[HTML]{0A7B41}} \color[HTML]{F1F1F1} 3.91 \\
&  &  & \ac{WIIG} & {\cellcolor[HTML]{EB5A3A}} \color[HTML]{F1F1F1} 0.83 ($\pm$0.15) & {\cellcolor[HTML]{B91326}} \color[HTML]{F1F1F1} 96.09 \\
 &  &  & \ac{VBIA} & {\cellcolor[HTML]{DC3B2C}} \color[HTML]{F1F1F1} \itshape 0.88 ($\pm$0.07) & {\cellcolor[HTML]{A70226}} \color[HTML]{F1F1F1} \itshape 99.22 \\
\cline{3-6}
&  & \multirow[c]{3}{*}{\ac{CVB}} & \ac{PIG} & {\cellcolor[HTML]{6BBF64}} \color[HTML]{000000} \bfseries 0.21 ($\pm$0.08) & {\cellcolor[HTML]{006837}} \color[HTML]{F1F1F1} \bfseries 0 \\
&  &  & \ac{WIIG} & {\cellcolor[HTML]{F67A49}} \color[HTML]{F1F1F1} 0.78 ($\pm$0.14) & {\cellcolor[HTML]{CA2427}} \color[HTML]{F1F1F1} 92.19 \\
 &  &  & \ac{VBIA} & {\cellcolor[HTML]{FBA05B}} \color[HTML]{000000} 0.72 ($\pm$0.13) & {\cellcolor[HTML]{C21C27}} \color[HTML]{F1F1F1} 93.75 \\
\cline{2-6}
& \multirow[c]{7}{*}{\ac{ViT}} & - & \ac{IG} & {\cellcolor[HTML]{D62F27}} \color[HTML]{F1F1F1} 0.90 ($\pm$0.05) & {\cellcolor[HTML]{A50026}} \color[HTML]{F1F1F1} \itshape 100 \\
\cline{3-6}
& & \multirow[c]{3}{*}{\ac{PRECODE}} & \ac{PIG} & {\cellcolor[HTML]{07753E}} \color[HTML]{F1F1F1} \bfseries 0.03 ($\pm$0.02) & {\cellcolor[HTML]{006837}} \color[HTML]{F1F1F1} \bfseries 0 \\
&  &  & \ac{WIIG} & {\cellcolor[HTML]{C62027}} \color[HTML]{F1F1F1} 0.93 ($\pm$0.03) & {\cellcolor[HTML]{A50026}} \color[HTML]{F1F1F1} \itshape 100 \\
 &  &  & \ac{VBIA} & {\cellcolor[HTML]{F16640}} \color[HTML]{F1F1F1} 0.81 ($\pm$0.07) & {\cellcolor[HTML]{A50026}} \color[HTML]{F1F1F1} \itshape 100 \\
\cline{3-6}
&  & \multirow[c]{3}{*}{\ac{CVB}} & \ac{PIG} & {\cellcolor[HTML]{45AD5B}} \color[HTML]{F1F1F1} 0.16 ($\pm$0.06) & {\cellcolor[HTML]{006837}} \color[HTML]{F1F1F1} \bfseries 0 \\
&  &  & \ac{WIIG} & {\cellcolor[HTML]{A50026}} \color[HTML]{F1F1F1} \itshape 1.00 ($\pm$0.00) & {\cellcolor[HTML]{A50026}} \color[HTML]{F1F1F1} \itshape 100 \\
 &  &  & \ac{VBIA} & {\cellcolor[HTML]{A50026}} \color[HTML]{F1F1F1} \itshape 1.00 ($\pm$0.00) & {\cellcolor[HTML]{A50026}} \color[HTML]{F1F1F1} \itshape 100 \\
\bottomrule
\end{tabular}
\end{table*}

The results of these experiments can be found in \autoref{tab:vbia_results_sdia}.
Although both \acp{PM} provide protection against the \ac{PIG} attack, their privacy preserving effect is significantly reduced when attacked with \ac{VBIA}. 
If the attacker has knowledge of the reparameterization matrix that was stochastically sampled during the client's training process as in the \ac{WIIG} attack, data can be reconstructed with high quality because the attacker uses the same model realization as the client, i.e., $\varepsilon_{\text{VB}_A} = \varepsilon_{\text{VB}_C}$.
However, even without this knowledge, the experiments show that the \ac{VBIA} attacker can sufficiently approximate the reparameterization matrix $\varepsilon_{\text{VB}_A} \approx \varepsilon_{\text{VB}_C}$. 
For the \ac{CNN}, the \ac{ASR} increases from less than $8\%$ with \ac{PIG} to over $93\%$ with \ac{VBIA} for both datasets and \acp{PM}. 
For the \ac{ViT}, the \ac{VBIA} achieves an \ac{ASR} of $100\%$ in all cases.

\subsection{Combined Defenses Inversion Attack}
The previous sections discussed \ac{GI} attacks to bypass the defensive mechanisms of Dropout (\ac{DIA} \parencite{scheliga2023dropout}), \ac{GP} (\ac{GPIA}), \ac{PRECODE}, and \ac{CVB} (\ac{VBIA}). 
We observed that if an attacker is able sufficiently approximate and mimic the client's gradient computation process, single defenses can be bypassed.
We therefore propose to combine multiple defense mechanisms.
This increases the complexity of the attack, as the attacker must approximate multiple sources of stochasticity to sufficiently mimic the client’s gradient computation process. 
This could lead to an unstable attack optimization process, which in turn would result in reduced reconstruction quality and higher privacy.
To find an optimal combination of defenses, we conduct an extensive ablation study that tests various combinations.

To rigorously test the defensive capabilities of these defense combinations, we formulate a \acf{CDIA}.
\ac{CDIA} combines the above attacks by targeting all applied defenses during the attack optimization process if the client used them during local training. 
Since \ac{DPIA} decreased reconstruction quality compared to \ac{IG}, we do not specifically target \ac{DP} in \ac{CDIA}.
For \ac{CDIA}, we assume that the attacker has knowledge of the defense mechanisms employed by the client.
This assumption is coherent with the threat model discussed in \autoref{sec:threat_model}. 
Dropout, \ac{PRECODE}, and \ac{CVB} are integrated into the model architecture and are known to the server, as it coordinates the federated training process. 
Even if the attacker does not directly possess this information, it could be inferred from the shapes of the transmitted gradients. 
While \ac{DP} might be more challenging to detect, \ac{CDIA} does not specifically target this defense. 
The use of \ac{GP}, however, can be easily inferred from the sparsity of the transmitted gradients.

\begin{algorithm}[t!]
\caption{\acf{CDIA}} 
\label{alg:CDIA}
\textbf{Input}: $F$: neural network; $\mathcal{L}$: training loss function; GDF: gradient distance function;\\
$\Tilde{G}=\text{\ac{GP}}_{\Phi_C}(\text{\ac{DP}}_{\Xi_C}(\nabla \mathcal{L}_W(F_{(\Psi_C, \varepsilon_{\text{VB}_C})}(x),y)))$: perturbed and pruned victim gradient;\\
$\eta_{\text{GI}}$: learning rate;\\
$\vdrop$: Dropout rate;\\
\textbf{Output}: $(x', y')$ training data reconstructions; \\
$\Psi_A = \lbrace\psi_A^{(1)},\dots\psi_A^{(\vLayerindex)}\rbrace$: approximated Dropout masks; \\
$\Phi_C$: client pruning mask; \\
$\varepsilon_{\text{VB}_A}$: approximated reparameterization matrix

\begin{algorithmic}[1]
    \State $x', y' \gets \mathcal{N}(0,\mathcal{I})$ \Comment{{\color{gray} initialize dummy data}}
    
    \If{Client used Dropout}
    \State $\psi_A^{(1)},\dots,\psi_A^{(\vLayerindex)} \gets \text{Bernoulli}(\vdrop)$ \Comment{{\color{gray} initialize dummy Dropout masks}}
    \EndIf
        
    \If{Client used \ac{GP}}
    \State $\Phi_C \gets \text{abs}(\Tilde{G})>0$ \Comment{{\color{gray} compute client pruning mask}}
    \EndIf

    \If{Client used \ac{PRECODE}/\ac{CVB}}
    \State $\varepsilon_{\text{VB}_A} \gets \mathcal{N}(0,\mathcal{I})$ \Comment{{\color{gray} initialize dummy reparameterization matrix}}
    \EndIf

    \While{not converged} \Comment{{\color{gray}reiterate until some termination criterion is reached}}
    \State $G' \gets \nabla \mathcal{L}_W(F_{(\Psi_A, \varepsilon_{\text{VB}_A})}(x'),y')$ \Comment{{\color{gray}calculate dummy gradient}}

        \If{Client used \ac{GP}}
        \State $\Tilde{G'} \gets \Tilde{G'} \odot \Phi_C$ \Comment{{\color{gray}apply pruning mask}}
        \EndIf
        
        \State $\mathrm{GD} \gets \mathrm{GDF}(\Tilde{G}, \Tilde{G'})$ \Comment{{\color{gray}calculate gradient distance}}
        \State $\mathrm{GD} \gets \mathrm{GD} + \vPrior$ \Comment{{\color{gray}add regularization terms}}
    \State $x' \gets x' - \eta_{\text{GI}} \frac{\partial \mathrm{GD}}{\partial x'}$; $y' \gets y' - \eta_{\text{GI}} \frac{\partial \mathrm{GD}}{\partial y'}$ \Comment{{\color{gray}update dummy data}}
        \If{Client used Dropout}
        \State $\psi_A^{(\vlayerindex)} \gets \psi_A^{(\vlayerindex)} - \eta_{\text{GI}} \frac{\partial \mathrm{GD}}{\partial \psi_A^{(\vlayerindex)}} \text{ } \forall \vlayerindex\in 1, \dots,\vLayerindex$ \Comment{{\color{gray}update dummy Dropout masks}}
        \EndIf    
        
        \If{Client used \ac{PRECODE}/\ac{CVB}}
        \State $\varepsilon_{\text{VB}_A} \gets \varepsilon_{\text{VB}_A} - \eta_{\text{GI}} \frac{\partial \mathrm{GD}}{\partial \varepsilon_{\text{VB}_A}}$ \Comment{{\color{gray}update dummy reparameterization matrix}}
        \EndIf		
    \EndWhile
  \State \textbf{return} $(x', y')$, $\Psi_A$, $\Phi_C$, $\varepsilon_{\text{VB}_A}$
\end{algorithmic}
\end{algorithm}

A detailed description of \ac{CDIA} is provided in \autoref{alg:CDIA}. 
The attack adapts based on the defenses used by the client. 
If Dropout was applied, \ac{CDIA} includes Dropout mask approximation in the joint optimization process. 
If \ac{GP} was used, the attacker calculates the ground truth pruning masks and applies them to the dummy gradient before calculating the gradient distance GD. 
For \ac{PRECODE} and \ac{CVB}, the attacker approximates the reparameterization matrix during optimization. 
For further details on each of the targeted attack mechanisms, please refer to the respective sections above.
Note, that when no defense is applied, \ac{CDIA} defaults to the basic \ac{IG} attack. 
When a single defense mechanism is used, \ac{CDIA} corresponds to \ac{DIA} for Dropout, \ac{IG} for \ac{DP}, \ac{GPIA} for \ac{GP}, and \ac{VBIA} for \ac{PRECODE} and \ac{CVB}. 

\section{Experimental Setup}
\label{sec:design}
This section describes the general settings for the experiments in this paper.
For most experiments we used the same simple \ac{CNN} model as in \parencite{scheliga2024privacy}.
It consists of three convolutional layers with $5\times5$ kernels, $\lbrack 16, 32, 64 \rbrack$ channels, a stride of $2$ and ReLU activation.
The last layer of the \ac{CNN} is a fully connected classification layer with 10 neurons and softmax activation.
The following sections describe more details on how we measured model utility and privacy leakage.

\subsection{Model Utility}
We train models for image classification on the MNIST~\parencite{lecun1998gradient} and CIFAR-10~\parencite{krizhevsky2009learning} datasets.
The datasets are first separated into training and test splits according to the corresponding benchmark protocols.
We embedded all experiments in a FL scenario with $10$ clients.
The training data splits are independent and identically distributed to those $10$ clients.
Each client creates a validation split that corresponds to $10\%$ of the training data.
This leaves every client with $5'400$/$4'500$ training samples, $600$/$400$ validation samples, and $1'000$/$1'000$ test samples for MNIST/CIFAR-10 respectively.

The clients collaboratively train a randomly initialized model for $300$ communication rounds with one local training epoch using \ac{FedAvg}~\parencite{mcmahan2017communication}.
Local training minimizes the cross-entropy loss using Adam optimizer~\parencite{kingma2014adam} with a learning rate or $0.001$, momentum parameters of $(\beta_1, \beta_2)=(0.9, 0.999)$, and a batch size of $64$.
To save computational resources, we stop the training early, if the mean validation loss over all clients has not improved for $40$ consecutive communication rounds.
To measure model utility, we calculate the accuracy of the global model states on the test data after every communication round.
We repeated each experiment and report the mean and standard deviation across three runs with different random seeds.

\subsection{Privacy}
For each of the datasets we create a victim dataset to evaluate privacy leakage.
The victim dataset is composed of $128$ images that are randomly sampled from the training data of one client.

We aim to empirically identify an upper bound for privacy leakage via iterative \ac{GI} attacks and test the limits of the investigated defense mechanisms.
To this end, we consider the \emph{worst-case defense scenario} for an attacked client.
Unless not otherwise stated, we use a batch size $\mathcal{B}=1$ and only perform a single local training step before transmitting the victim gradient to the attacker.
As discussed in~\autoref{sec:sota_def}, previous work has shown that higher batch size and multiple local training iterations increase the difficulty of the attack, thereby decreasing privacy leakage.
If the defense mechanism works even in such \emph{hard to defend} settings, privacy leakage becomes even more improbable in realistic training scenarios~\parencite{wei2020framework, huang2021evaluating}.

We use our proposed \ac{CDIA} attack to target and bypass each of applied defenses.
We configure it so that it defaults to a baseline \ac{IG} attack~\parencite{geiping2020inverting} if no defenses or only \ac{DP} is applied.
In detail, dummy images are initialized from a Gaussian distribution, cosine distance is used as loss function, and \ac{TV} as regularization term with weight $\lambda_{\text{TV}}=0.01$.
We use Adam optimizer with initial learning rate $1$  and reduce it by a factor of $0.1$ if the reconstruction loss plateaus for $400$ attack iterations.
To save computational resources, attacks are stopped if either the reconstruction loss falls below a value of $10^{-5}$, or there is no decrease in reconstruction loss for $4'000$ iterations, or after a maximum of $20,000$ iterations.
Furthermore, we assume that the label information for each reconstruction is known, as it can be analytically reconstructed from gradients of cross-entropy loss functions with respect to the weights of fully connected layers with softmax activation~\parencite{geiping2020inverting, zhao2020idlg, wei2020framework}.
We ensured that our threat model (\cf \autoref{sec:threat_model}) and the described attack simulation setup is consistent with related work~\parencite{zhu2019deep, geiping2020inverting, wei2020framework, scheliga2023dropout, scheliga2024privacy}.

For each reconstructed image, we calculate the reconstruction quality by comparing it with the original image from the victim dataset. 
While \ac{MSE} and \ac{PSNR} are commonly used metrics for assessing reconstruction quality, they often do not correlate well with human visual perception~\parencite{wang2004image}. 
Therefore, we use \ac{SSIM}~\parencite{wang2004image} as the primary metric for measuring reconstruction quality. 
Since we report the average \ac{SSIM} value over all reconstructions, extremely good or bad reconstructions might be missed.
Therefore, we additionally compute the \ac{ASR} as the fraction of the successfully reconstructed images of the victim dataset~\parencite{wagner2018technical}.
We consider an image to be successfully reconstructed if the \ac{SSIM} value is above a threshold of $\vthreshASR = 0.5$.

\subsection{Ablation Study of Defense Mechanisms}
We conduct an extensive ablation study, to examine all possible combinations of the previously discussed defense mechanisms, including Dropout, \ac{DP}, \ac{GP}, and one of the \acp{PM} (either \ac{PRECODE} or \ac{CVB}). 
For Dropout, we apply a modest Dropout rate of $\vdrop=0.25$, as previous studies showed a decrease of model utility for higher rates~\parencite{scheliga2023dropout}. 
When using perturbation-based defenses, we maintain low levels of perturbation to preserve an acceptable trade-off between model utility and privacy. 
Specifically, we use a clipping threshold of $\vthreshclip = 20$ and a noise multiplier $\sigma = 10^{-4}$ for \ac{DP}, and a pruning rate of $\vprune = 0.50$ for \ac{GP}.
Previous studies indicate that higher perturbation rates have a significant negative impact on model utility~\parencite{scheliga2024privacy}.
For \ac{PRECODE} and \ac{CVB} we apply the hyperparameter configurations recommended in~\parencite{scheliga2024privacy}.

\section{Experimental Results \& Discussion}
\subsection{Ablation Study}
\begin{figure}
    \centering
    \includegraphics[width=0.8\linewidth]{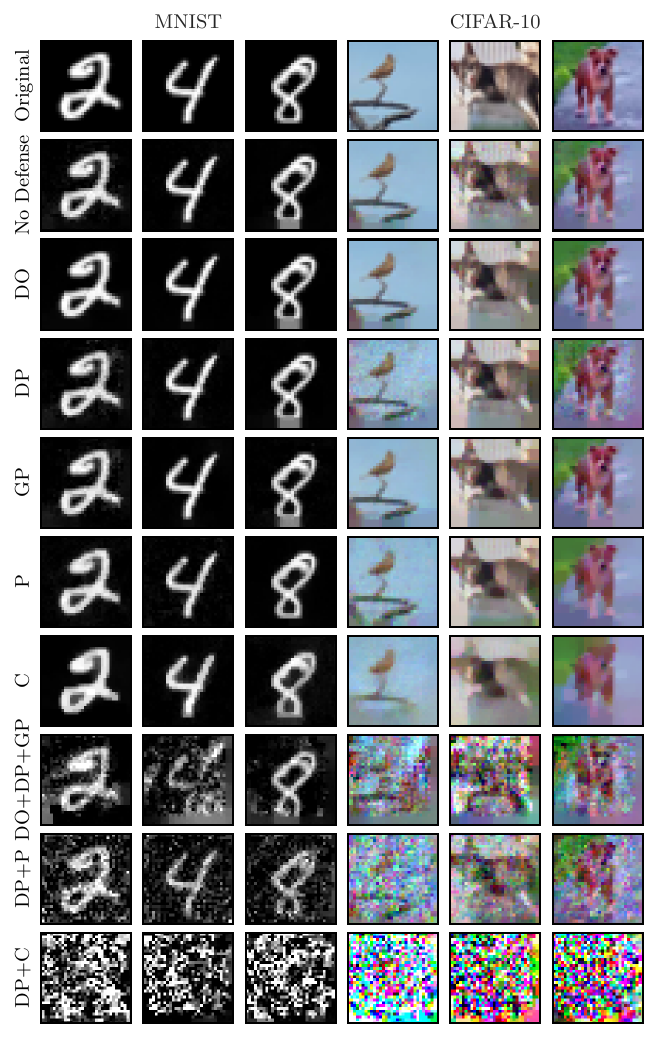}
    \caption{
    Example reconstructions for the \ac{CNN} on the MNIST and CIFAR-10 datasets when protected with different defense combinations.
    We illustrate examples for Dropout (DO), \ac{DP}, \ac{GP}, \ac{PRECODE} (P) and \ac{CVB} (C).
    }
    \label{fig:ablation}
\end{figure}

First, we conducted an extensive ablation study that investigates the impact on model utility and privacy of all potential defense combinations of the investigated defenses.
The results for these experiments can be found in~\autoref{tab:results_xcomb_ablation_mnist}.
Example reconstructions are displayed in~\autoref{fig:ablation}.
\begin{table*}[h!]
\centering
\caption{
Model utility and privacy metrics for an ablation study using different combinations of defense mechanisms in the \ac{CNN} on the MNIST and CIFAR-10 dataset. 
We consider Dropout (DO), \acf{DP}, \acf{GP} and a \acf{PM} as defense.
\cmark and \textcolor[HTML]{cfcfcf}{\xmark} ~indicate, whether a defense mechanism was used or not.
P and C indicate whether \ac{PRECODE} or \ac{CVB} was used, respectively.
Arrows indicate direction of improvement from the viewpoint of a defending client.
Bold and italic formatting highlight best and worst results, respectively.
}
\label{tab:results_xcomb_ablation_mnist}
\resizebox{0.7\linewidth}{!}{%
\begin{tabular}{c|c|c|c|c|c|c|c}
\toprule
 & DO & \ac{DP} & \ac{GP} & PM & \ac{SSIM} $\downarrow$ & \ac{ASR} [\%] $\downarrow$ & Accuracy [\%] $\uparrow$ \\
\midrule
\multirow[c]{24}{*}{\rotatebox{90}{MNIST}} & \color[HTML]{cfcfcf} \xmark  & \color[HTML]{cfcfcf} \xmark  & \color[HTML]{cfcfcf} \xmark  & \color[HTML]{cfcfcf} \xmark  & {\cellcolor[HTML]{BD1726}} \color[HTML]{F1F1F1} 0.95 ($\pm$0.06) & {\cellcolor[HTML]{A50026}} \color[HTML]{F1F1F1} \itshape 100 & {\cellcolor[HTML]{006837}} \color[HTML]{F1F1F1} 99.10 ($\pm$0.04) \\
 & \color[HTML]{cfcfcf} \xmark  & \color[HTML]{cfcfcf} \xmark  & \color[HTML]{cfcfcf} \xmark  & P & {\cellcolor[HTML]{BD1726}} \color[HTML]{F1F1F1} 0.95 ($\pm$0.07) & {\cellcolor[HTML]{A50026}} \color[HTML]{F1F1F1} \itshape 100 & {\cellcolor[HTML]{026C39}} \color[HTML]{F1F1F1} 98.33 ($\pm$0.02) \\
 & \color[HTML]{cfcfcf} \xmark  & \color[HTML]{cfcfcf} \xmark  & \color[HTML]{cfcfcf} \xmark  & C & {\cellcolor[HTML]{DC3B2C}} \color[HTML]{F1F1F1} 0.88 ($\pm$0.09) & {\cellcolor[HTML]{A70226}} \color[HTML]{F1F1F1} 99.22 & {\cellcolor[HTML]{006837}} \color[HTML]{F1F1F1} 99.15 ($\pm$0.05) \\
 & \color[HTML]{cfcfcf} \xmark  & \color[HTML]{cfcfcf} \xmark  & \cmark & \color[HTML]{cfcfcf} \xmark  & {\cellcolor[HTML]{BD1726}} \color[HTML]{F1F1F1} 0.95 ($\pm$0.04) & {\cellcolor[HTML]{A50026}} \color[HTML]{F1F1F1} \itshape 100 & {\cellcolor[HTML]{006837}} \color[HTML]{F1F1F1} 99.00 ($\pm$0.09) \\
 & \color[HTML]{cfcfcf} \xmark  & \color[HTML]{cfcfcf} \xmark  & \cmark & P & {\cellcolor[HTML]{C62027}} \color[HTML]{F1F1F1} 0.93 ($\pm$0.09) & {\cellcolor[HTML]{A50026}} \color[HTML]{F1F1F1} \itshape 100 & {\cellcolor[HTML]{026C39}} \color[HTML]{F1F1F1} 98.41 ($\pm$0.04) \\
 & \color[HTML]{cfcfcf} \xmark  & \color[HTML]{cfcfcf} \xmark  & \cmark & C & {\cellcolor[HTML]{E75337}} \color[HTML]{F1F1F1} 0.84 ($\pm$0.08) & {\cellcolor[HTML]{A70226}} \color[HTML]{F1F1F1} 99.22 & {\cellcolor[HTML]{006837}} \color[HTML]{F1F1F1} 99.06 ($\pm$0.05) \\
 & \color[HTML]{cfcfcf} \xmark  & \cmark & \color[HTML]{cfcfcf} \xmark  & \color[HTML]{cfcfcf} \xmark  & {\cellcolor[HTML]{C62027}} \color[HTML]{F1F1F1} 0.93 ($\pm$0.06) & {\cellcolor[HTML]{A50026}} \color[HTML]{F1F1F1} \itshape 100 & {\cellcolor[HTML]{006837}} \color[HTML]{F1F1F1} 99.06 ($\pm$0.00) \\
 & \color[HTML]{cfcfcf} \xmark  & \cmark & \color[HTML]{cfcfcf} \xmark  & P & {\cellcolor[HTML]{FBFDBA}} \color[HTML]{000000} 0.49 ($\pm$0.12) & {\cellcolor[HTML]{FFFDBC}} \color[HTML]{000000} 50.78 & {\cellcolor[HTML]{026C39}} \color[HTML]{F1F1F1} \itshape 98.19 ($\pm$0.03) \\
 & \color[HTML]{cfcfcf} \xmark  & \cmark & \color[HTML]{cfcfcf} \xmark  & C & {\cellcolor[HTML]{36A657}} \color[HTML]{F1F1F1} 0.14 ($\pm$0.05) & {\cellcolor[HTML]{006837}} \color[HTML]{F1F1F1} \bfseries 0 & {\cellcolor[HTML]{006837}} \color[HTML]{F1F1F1} 99.18 ($\pm$0.00) \\
 & \color[HTML]{cfcfcf} \xmark  & \cmark & \cmark & \color[HTML]{cfcfcf} \xmark  & {\cellcolor[HTML]{BD1726}} \color[HTML]{F1F1F1} 0.95 ($\pm$0.05) & {\cellcolor[HTML]{A50026}} \color[HTML]{F1F1F1} \itshape 100 & {\cellcolor[HTML]{006837}} \color[HTML]{F1F1F1} 98.97 ($\pm$0.05) \\
 & \color[HTML]{cfcfcf} \xmark  & \cmark & \cmark & P & {\cellcolor[HTML]{FFFEBE}} \color[HTML]{000000} 0.50 ($\pm$0.13) & {\cellcolor[HTML]{FFF0A6}} \color[HTML]{000000} 54.69 & {\cellcolor[HTML]{026C39}} \color[HTML]{F1F1F1} 98.22 ($\pm$0.14) \\
 & \color[HTML]{cfcfcf} \xmark  & \cmark & \cmark & C & {\cellcolor[HTML]{36A657}} \color[HTML]{F1F1F1} 0.14 ($\pm$0.04) & {\cellcolor[HTML]{006837}} \color[HTML]{F1F1F1} \bfseries 0 & {\cellcolor[HTML]{006837}} \color[HTML]{F1F1F1} \bfseries 99.19 ($\pm$0.05) \\
 & \cmark & \color[HTML]{cfcfcf} \xmark  & \color[HTML]{cfcfcf} \xmark  & \color[HTML]{cfcfcf} \xmark  & {\cellcolor[HTML]{B30D26}} \color[HTML]{F1F1F1} \itshape 0.97 ($\pm$0.02) & {\cellcolor[HTML]{A50026}} \color[HTML]{F1F1F1} \itshape 100 & {\cellcolor[HTML]{006837}} \color[HTML]{F1F1F1} 99.12 ($\pm$0.03) \\
 & \cmark & \color[HTML]{cfcfcf} \xmark  & \color[HTML]{cfcfcf} \xmark  & P & {\cellcolor[HTML]{B30D26}} \color[HTML]{F1F1F1} \itshape 0.97 ($\pm$0.04) & {\cellcolor[HTML]{A50026}} \color[HTML]{F1F1F1} \itshape 100 & {\cellcolor[HTML]{026C39}} \color[HTML]{F1F1F1} 98.41 ($\pm$0.02) \\
 & \cmark & \color[HTML]{cfcfcf} \xmark  & \color[HTML]{cfcfcf} \xmark  & C & {\cellcolor[HTML]{D22B27}} \color[HTML]{F1F1F1} 0.91 ($\pm$0.06) & {\cellcolor[HTML]{A50026}} \color[HTML]{F1F1F1} \itshape 100 & {\cellcolor[HTML]{006837}} \color[HTML]{F1F1F1} 99.09 ($\pm$0.08) \\
 & \cmark & \color[HTML]{cfcfcf} \xmark  & \cmark & \color[HTML]{cfcfcf} \xmark  & {\cellcolor[HTML]{B30D26}} \color[HTML]{F1F1F1} \itshape 0.97 ($\pm$0.02) & {\cellcolor[HTML]{A50026}} \color[HTML]{F1F1F1} \itshape 100 & {\cellcolor[HTML]{006837}} \color[HTML]{F1F1F1} 98.95 ($\pm$0.06) \\
 & \cmark & \color[HTML]{cfcfcf} \xmark  & \cmark & P & {\cellcolor[HTML]{B91326}} \color[HTML]{F1F1F1} 0.96 ($\pm$0.06) & {\cellcolor[HTML]{A50026}} \color[HTML]{F1F1F1} \itshape 100 & {\cellcolor[HTML]{026C39}} \color[HTML]{F1F1F1} 98.24 ($\pm$0.03) \\
 & \cmark & \color[HTML]{cfcfcf} \xmark  & \cmark & C & {\cellcolor[HTML]{E54E35}} \color[HTML]{F1F1F1} 0.85 ($\pm$0.09) & {\cellcolor[HTML]{A70226}} \color[HTML]{F1F1F1} 99.22 & {\cellcolor[HTML]{006837}} \color[HTML]{F1F1F1} 99.05 ($\pm$0.03) \\
 & \cmark & \cmark & \color[HTML]{cfcfcf} \xmark  & \color[HTML]{cfcfcf} \xmark  & {\cellcolor[HTML]{FEEC9F}} \color[HTML]{000000} 0.56 ($\pm$0.14) & {\cellcolor[HTML]{FDB163}} \color[HTML]{000000} 69.53 & {\cellcolor[HTML]{006837}} \color[HTML]{F1F1F1} 99.06 ($\pm$0.05) \\
 & \cmark & \cmark & \color[HTML]{cfcfcf} \xmark  & P & {\cellcolor[HTML]{D9EF8B}} \color[HTML]{000000} 0.40 ($\pm$0.20) & {\cellcolor[HTML]{D3EC87}} \color[HTML]{000000} 39.06 & {\cellcolor[HTML]{026C39}} \color[HTML]{F1F1F1} 98.28 ($\pm$0.00) \\
 & \cmark & \cmark & \color[HTML]{cfcfcf} \xmark  & C & {\cellcolor[HTML]{279F53}} \color[HTML]{F1F1F1} \bfseries 0.12 ($\pm$0.06) & {\cellcolor[HTML]{006837}} \color[HTML]{F1F1F1} \bfseries 0 & {\cellcolor[HTML]{006837}} \color[HTML]{F1F1F1} 99.12 ($\pm$0.01) \\
 & \cmark & \cmark & \cmark & \color[HTML]{cfcfcf} \xmark  & {\cellcolor[HTML]{FEEC9F}} \color[HTML]{000000} 0.56 ($\pm$0.14) & {\cellcolor[HTML]{FBA05B}} \color[HTML]{000000} 71.88 & {\cellcolor[HTML]{006837}} \color[HTML]{F1F1F1} 98.95 ($\pm$0.04) \\
 & \cmark & \cmark & \cmark & P & {\cellcolor[HTML]{C9E881}} \color[HTML]{000000} 0.37 ($\pm$0.19) & {\cellcolor[HTML]{93D168}} \color[HTML]{000000} 27.34 & {\cellcolor[HTML]{026C39}} \color[HTML]{F1F1F1} 98.23 ($\pm$0.03) \\
 & \cmark & \cmark & \cmark & C & {\cellcolor[HTML]{279F53}} \color[HTML]{F1F1F1} \bfseries 0.12 ($\pm$0.05) & {\cellcolor[HTML]{006837}} \color[HTML]{F1F1F1} \bfseries 0 & {\cellcolor[HTML]{006837}} \color[HTML]{F1F1F1} 99.15 ($\pm$0.06) \\

\midrule

\multirow[c]{24}{*}{\rotatebox{90}{CIFAR-10}} & \color[HTML]{cfcfcf} \xmark  & \color[HTML]{cfcfcf} \xmark  & \color[HTML]{cfcfcf} \xmark  & \color[HTML]{cfcfcf} \xmark  & {\cellcolor[HTML]{E0422F}} \color[HTML]{F1F1F1} 0.87 ($\pm$0.11) & {\cellcolor[HTML]{B30D26}} \color[HTML]{F1F1F1} 96.88 & {\cellcolor[HTML]{17934E}} \color[HTML]{F1F1F1} 62.53 ($\pm$0.16) \\
 & \color[HTML]{cfcfcf} \xmark  & \color[HTML]{cfcfcf} \xmark  & \color[HTML]{cfcfcf} \xmark  & P & {\cellcolor[HTML]{DC3B2C}} \color[HTML]{F1F1F1} 0.88 ($\pm$0.07) & {\cellcolor[HTML]{A70226}} \color[HTML]{F1F1F1} 99.22 & {\cellcolor[HTML]{39A758}} \color[HTML]{F1F1F1} 59.48 ($\pm$0.23) \\
 & \color[HTML]{cfcfcf} \xmark  & \color[HTML]{cfcfcf} \xmark  & \color[HTML]{cfcfcf} \xmark  & C & {\cellcolor[HTML]{FBA05B}} \color[HTML]{000000} 0.72 ($\pm$0.13) & {\cellcolor[HTML]{C21C27}} \color[HTML]{F1F1F1} 93.75 & {\cellcolor[HTML]{026C39}} \color[HTML]{F1F1F1} 67.19 ($\pm$0.64) \\
 & \color[HTML]{cfcfcf} \xmark  & \color[HTML]{cfcfcf} \xmark  & \cmark & \color[HTML]{cfcfcf} \xmark  & {\cellcolor[HTML]{E24731}} \color[HTML]{F1F1F1} 0.86 ($\pm$0.09) & {\cellcolor[HTML]{AB0626}} \color[HTML]{F1F1F1} 98.44 & {\cellcolor[HTML]{17934E}} \color[HTML]{F1F1F1} 62.53 ($\pm$0.34) \\
 & \color[HTML]{cfcfcf} \xmark  & \color[HTML]{cfcfcf} \xmark  & \cmark & P & {\cellcolor[HTML]{E24731}} \color[HTML]{F1F1F1} 0.86 ($\pm$0.10) & {\cellcolor[HTML]{AB0626}} \color[HTML]{F1F1F1} 98.44 & {\cellcolor[HTML]{39A758}} \color[HTML]{F1F1F1} 59.53 ($\pm$0.11) \\
 & \color[HTML]{cfcfcf} \xmark  & \color[HTML]{cfcfcf} \xmark  & \cmark & C & {\cellcolor[HTML]{FDC372}} \color[HTML]{000000} 0.66 ($\pm$0.09) & {\cellcolor[HTML]{C62027}} \color[HTML]{F1F1F1} 92.97 & {\cellcolor[HTML]{026C39}} \color[HTML]{F1F1F1} 67.34 ($\pm$0.28) \\
 & \color[HTML]{cfcfcf} \xmark  & \cmark & \color[HTML]{cfcfcf} \xmark  & \color[HTML]{cfcfcf} \xmark  & {\cellcolor[HTML]{EB5A3A}} \color[HTML]{F1F1F1} 0.83 ($\pm$0.11) & {\cellcolor[HTML]{AF0926}} \color[HTML]{F1F1F1} 97.66 & {\cellcolor[HTML]{18954F}} \color[HTML]{F1F1F1} 62.31 ($\pm$0.00) \\
 & \color[HTML]{cfcfcf} \xmark  & \cmark & \color[HTML]{cfcfcf} \xmark  & P & {\cellcolor[HTML]{D9EF8B}} \color[HTML]{000000} 0.40 ($\pm$0.14) & {\cellcolor[HTML]{89CC67}} \color[HTML]{000000} 25.78 & {\cellcolor[HTML]{57B65F}} \color[HTML]{F1F1F1} 57.39 ($\pm$0.26) \\
 & \color[HTML]{cfcfcf} \xmark  & \cmark & \color[HTML]{cfcfcf} \xmark  & C & {\cellcolor[HTML]{0A7B41}} \color[HTML]{F1F1F1} \bfseries 0.04 ($\pm$0.02) & {\cellcolor[HTML]{006837}} \color[HTML]{F1F1F1} \bfseries 0 & {\cellcolor[HTML]{08773F}} \color[HTML]{F1F1F1} 65.86 ($\pm$0.25) \\
 & \color[HTML]{cfcfcf} \xmark  & \cmark & \cmark & \color[HTML]{cfcfcf} \xmark  & {\cellcolor[HTML]{F16640}} \color[HTML]{F1F1F1} 0.81 ($\pm$0.10) & {\cellcolor[HTML]{B30D26}} \color[HTML]{F1F1F1} 96.88 & {\cellcolor[HTML]{2AA054}} \color[HTML]{F1F1F1} 60.72 ($\pm$0.51) \\
 & \color[HTML]{cfcfcf} \xmark  & \cmark & \cmark & P & {\cellcolor[HTML]{BFE47A}} \color[HTML]{000000} 0.35 ($\pm$0.14) & {\cellcolor[HTML]{4BB05C}} \color[HTML]{F1F1F1} 16.41 & {\cellcolor[HTML]{66BD63}} \color[HTML]{F1F1F1} \itshape 56.26 ($\pm$0.01) \\
 & \color[HTML]{cfcfcf} \xmark  & \cmark & \cmark & C & {\cellcolor[HTML]{0A7B41}} \color[HTML]{F1F1F1} \bfseries 0.04 ($\pm$0.02) & {\cellcolor[HTML]{006837}} \color[HTML]{F1F1F1} \bfseries 0 & {\cellcolor[HTML]{0A7B41}} \color[HTML]{F1F1F1} 65.45 ($\pm$1.04) \\
 & \cmark & \color[HTML]{cfcfcf} \xmark  & \color[HTML]{cfcfcf} \xmark  & \color[HTML]{cfcfcf} \xmark  & {\cellcolor[HTML]{D22B27}} \color[HTML]{F1F1F1} \itshape 0.91 ($\pm$0.05) & {\cellcolor[HTML]{A50026}} \color[HTML]{F1F1F1} \itshape 100 & {\cellcolor[HTML]{0F8446}} \color[HTML]{F1F1F1} 64.36 ($\pm$0.12) \\
 & \cmark & \color[HTML]{cfcfcf} \xmark  & \color[HTML]{cfcfcf} \xmark  & P & {\cellcolor[HTML]{DA362A}} \color[HTML]{F1F1F1} 0.89 ($\pm$0.07) & {\cellcolor[HTML]{A70226}} \color[HTML]{F1F1F1} 99.22 & {\cellcolor[HTML]{36A657}} \color[HTML]{F1F1F1} 59.77 ($\pm$0.16) \\
 & \cmark & \color[HTML]{cfcfcf} \xmark  & \color[HTML]{cfcfcf} \xmark  & C & {\cellcolor[HTML]{F88C51}} \color[HTML]{F1F1F1} 0.75 ($\pm$0.08) & {\cellcolor[HTML]{A50026}} \color[HTML]{F1F1F1} \itshape 100 & {\cellcolor[HTML]{006837}} \color[HTML]{F1F1F1} \bfseries 67.80 ($\pm$0.41) \\
 & \cmark & \color[HTML]{cfcfcf} \xmark  & \cmark & \color[HTML]{cfcfcf} \xmark  & {\cellcolor[HTML]{DC3B2C}} \color[HTML]{F1F1F1} 0.88 ($\pm$0.05) & {\cellcolor[HTML]{A50026}} \color[HTML]{F1F1F1} \itshape 100 & {\cellcolor[HTML]{17934E}} \color[HTML]{F1F1F1} 62.43 ($\pm$0.24) \\
 & \cmark & \color[HTML]{cfcfcf} \xmark  & \cmark & P & {\cellcolor[HTML]{E0422F}} \color[HTML]{F1F1F1} 0.87 ($\pm$0.07) & {\cellcolor[HTML]{A50026}} \color[HTML]{F1F1F1} \itshape 100 & {\cellcolor[HTML]{36A657}} \color[HTML]{F1F1F1} 59.83 ($\pm$0.34) \\
 & \cmark & \color[HTML]{cfcfcf} \xmark  & \cmark & C & {\cellcolor[HTML]{FDBD6D}} \color[HTML]{000000} 0.67 ($\pm$0.11) & {\cellcolor[HTML]{BD1726}} \color[HTML]{F1F1F1} 95.31 & {\cellcolor[HTML]{006837}} \color[HTML]{F1F1F1} 67.68 ($\pm$0.48) \\
 & \cmark & \cmark & \color[HTML]{cfcfcf} \xmark  & \color[HTML]{cfcfcf} \xmark  & {\cellcolor[HTML]{C9E881}} \color[HTML]{000000} 0.37 ($\pm$0.14) & {\cellcolor[HTML]{60BA62}} \color[HTML]{F1F1F1} 19.53 & {\cellcolor[HTML]{148E4B}} \color[HTML]{F1F1F1} 63.21 ($\pm$0.27) \\
 & \cmark & \cmark & \color[HTML]{cfcfcf} \xmark  & P & {\cellcolor[HTML]{ABDB6D}} \color[HTML]{000000} 0.31 ($\pm$0.17) & {\cellcolor[HTML]{3CA959}} \color[HTML]{F1F1F1} 14.84 & {\cellcolor[HTML]{57B65F}} \color[HTML]{F1F1F1} 57.32 ($\pm$0.19) \\
 & \cmark & \cmark & \color[HTML]{cfcfcf} \xmark  & C & {\cellcolor[HTML]{0A7B41}} \color[HTML]{F1F1F1} \bfseries 0.04 ($\pm$0.02) & {\cellcolor[HTML]{006837}} \color[HTML]{F1F1F1} \bfseries 0 & {\cellcolor[HTML]{07753E}} \color[HTML]{F1F1F1} 66.00 ($\pm$0.42) \\
 & \cmark & \cmark & \cmark & \color[HTML]{cfcfcf} \xmark  & {\cellcolor[HTML]{C9E881}} \color[HTML]{000000} 0.37 ($\pm$0.13) & {\cellcolor[HTML]{4BB05C}} \color[HTML]{F1F1F1} 16.41 & {\cellcolor[HTML]{1E9A51}} \color[HTML]{F1F1F1} 61.51 ($\pm$0.14) \\
 & \cmark & \cmark & \cmark & P & {\cellcolor[HTML]{87CB67}} \color[HTML]{000000} 0.25 ($\pm$0.17) & {\cellcolor[HTML]{108647}} \color[HTML]{F1F1F1} 6.25 & {\cellcolor[HTML]{66BD63}} \color[HTML]{F1F1F1} \itshape 56.26 ($\pm$0.01) \\
 & \cmark & \cmark & \cmark & C & {\cellcolor[HTML]{0A7B41}} \color[HTML]{F1F1F1} \bfseries 0.04 ($\pm$0.02) & {\cellcolor[HTML]{006837}} \color[HTML]{F1F1F1} \bfseries 0 & {\cellcolor[HTML]{0A7B41}} \color[HTML]{F1F1F1} 65.38 ($\pm$1.01) \\
\bottomrule
\end{tabular} %
}
\end{table*}

For the MNIST dataset, model utility did not vary significantly across the defense combinations. 
The baseline model without any defense achieved an accuracy of $99.10\%$. 
Models utilizing a \ac{CVB} generally performed slightly better, with the highest accuracy of $99.19\%$ achieved by the \ac{DP}+\ac{GP}+\ac{CVB} combination. 
However, most other defense combinations slightly reduced model utility, with the lowest accuracy of $98.19\%$ resulting from the \ac{DP}+\ac{PRECODE} combination.
When attacked by \ac{CDIA}, single defenses and most defense combinations were unable to protect the privacy of the training data, as indicated by an \ac{ASR} close to $100\%$. 
Without a \ac{PM}, the lowest \ac{ASR} achieved was around $70\%$ for DO+\ac{DP} and DO+\ac{DP}+\ac{GP}. 
However, defense combinations involving \ac{PRECODE} reduced \ac{ASR} to between $27.34\%$ and $54.69\%$ for DO+\ac{DP}+\ac{GP}+\ac{PRECODE} and \ac{DP}+\ac{GP}+\ac{PRECODE}, respectively. 
While these combinations improved privacy to some extent, they also reduced model utility compared to the baseline.
Only defense combinations utilizing both \ac{DP} and \ac{CVB} reduced \ac{ASR} to $0\%$. 
The gradient scaling and slight noise introduced by \ac{DP} appears sufficient to disrupt \ac{CDIA}, specifically hindering the optimization of $\varepsilon_{\text{VB}_A}$ and destabilizing the attack optimization. 
For MNIST, the best combination in terms of both model utility and privacy was \ac{DP}+\ac{GP}+\ac{CVB}, resulting in an \ac{ASR} of $0\%$ and an accuracy of $99.19\%$.

For the CIFAR-10 dataset, both accuracy and reconstruction quality were consistently lower due to the higher complexity of the dataset.
However, the general trends for the defense combinations remained similar. 
Defense combinations involving \ac{CVB} noticeably improved accuracy compared to the unprotected baseline model, with the highest accuracy of $67.80\%$ achieved by the DO+\ac{CVB} combination, representing an improvement of $5.27\%$. 
Introducing \ac{PRECODE} harmed model utility more significantly on CIFAR-10, with the worst case being \ac{DP}+\ac{GP}+\ac{PRECODE}, where accuracy dropped by $6.27\%$ compared to the baseline model.
Again, single defenses and most defense combinations were ineffective at protecting privacy, with \ac{ASR} remaining close to $100\%$.
Combinations involving DO+\ac{DP}, \ac{DP}+\ac{PRECODE}, and \ac{DP}+\ac{CVB} were the only ones that significantly reduced \ac{ASR} to below $20\%$. 
Notably, only the combinations of \ac{DP} and \ac{CVB} consistently reduced \ac{ASR} to $0\%$, highlighting the importance of this combination for privacy protection against \ac{CDIA}.

While DO and \ac{GP} are not strictly required for privacy preservation, they can enhance model utility due to their regularizing effects, provided they are applied with moderate Dropout and pruning rates. 
For CIFAR-10, the best combination in terms of both model utility and privacy was DO+\ac{DP}+\ac{CVB}, which resulted in an \ac{ASR} of $0\%$ and an accuracy of $66\%$, boosting model utility by $3.47\%$ compared to the unprotected baseline model.

\subsection{Application to other Model Architectures}
Next, we investigate the trade-off between model utility and privacy when applying various defense mechanism combinations to different model architectures. 
Specifically, we consider a ResNet-18~\parencite{he2016deep} and a small \ac{ViT}~\parencite{dosovitskiy2020image} as they implement seminal architectural mechanisms.
The \ac{ViT} uses an embedding patch size 4, 4 transformer blocks with a hidden size 256, 16 attention heads and GELU activation.
The ResNet-18 uses the pre-defined architecture as described in the original paper~\parencite{he2016deep}.
For both models the final classification layer uses 10 neurons with softmax activation.
Additionally, we replace \ac{BN} layers with \ac{GN}~\parencite{wu2018group} as these are the preferred choice in \ac{FL} scenarios~\parencite{hsieh2020non, zhang2021improving}.
To save computational resources, we focus only on the most promising defense combinations identified in the previous experiments.
We include experiments for single defenses for reference.
Since \ac{GP} did not show significant benefits for either privacy or model utility, we exclude it from further consideration. 

\begin{table*}[h!]
\centering
\caption{
Model utility and privacy metrics for an ablation study using different combinations of defense mechanisms in the ResNet-18 and \ac{ViT} on the MNIST and CIFAR-10 datasets. 
We consider Dropout (DO), \acf{DP}, \acf{GP} and a \acf{PM} as defense.
\cmark and \textcolor[HTML]{cfcfcf}{\xmark} indicate, whether a defense mechanism was used or not.
P and C indicate whether \ac{PRECODE} or \ac{CVB} was used, respectively.
Arrows indicate direction of improvement from the viewpoint of a defending client.
Bold and italic formatting highlight best and worst results, respectively.
}
\label{tab:results_xcomb_models}
\resizebox{0.9\textwidth}{!}{%

\begin{tabular}{c|c|c|c|c|c|c|c|c}
\toprule
& Model & DO & \ac{DP} & \ac{GP} & PM & \ac{SSIM} $\downarrow$ & \ac{ASR} [\%] $\downarrow$ & Accuracy [\%] $\uparrow$ \\
\midrule
\multirow[c]{18}{*}{\rotatebox{90}{MNIST}} & \multirow[c]{9}{*}{ResNet-18} & \color[HTML]{cfcfcf} \xmark & \color[HTML]{cfcfcf} \xmark & \color[HTML]{cfcfcf} \xmark & \color[HTML]{cfcfcf} \xmark & {\cellcolor[HTML]{FFFCBA}} \color[HTML]{000000} 0.51 ($\pm$0.12) & {\cellcolor[HTML]{FFF2AA}} \color[HTML]{000000} 53.91 & {\cellcolor[HTML]{006837}} \color[HTML]{F1F1F1} 99.51 ($\pm$0.02) \\
 &  & \color[HTML]{cfcfcf} \xmark & \color[HTML]{cfcfcf} \xmark & \color[HTML]{cfcfcf} \xmark & P & {\cellcolor[HTML]{87CB67}} \color[HTML]{000000} 0.25 ($\pm$0.15) & {\cellcolor[HTML]{118848}} \color[HTML]{F1F1F1} 7.03 & {\cellcolor[HTML]{006837}} \color[HTML]{F1F1F1} \itshape 99.35 ($\pm$0.04) \\
 &  & \color[HTML]{cfcfcf} \xmark & \color[HTML]{cfcfcf} \xmark & \color[HTML]{cfcfcf} \xmark & C & {\cellcolor[HTML]{C9E881}} \color[HTML]{000000} 0.37 ($\pm$0.12) & {\cellcolor[HTML]{36A657}} \color[HTML]{F1F1F1} 14.06 & {\cellcolor[HTML]{006837}} \color[HTML]{F1F1F1} 99.46 ($\pm$0.01) \\
 &  & \color[HTML]{cfcfcf} \xmark & \cmark & \color[HTML]{cfcfcf} \xmark & \color[HTML]{cfcfcf} \xmark & {\cellcolor[HTML]{EFF8AA}} \color[HTML]{000000} 0.46 ($\pm$0.11) & {\cellcolor[HTML]{D3EC87}} \color[HTML]{000000} 39.06 & {\cellcolor[HTML]{006837}} \color[HTML]{F1F1F1} 99.52 ($\pm$0.04) \\
 &  & \color[HTML]{cfcfcf} \xmark & \cmark & \color[HTML]{cfcfcf} \xmark & P & {\cellcolor[HTML]{6BBF64}} \color[HTML]{000000} 0.21 ($\pm$0.11) & {\cellcolor[HTML]{006837}} \color[HTML]{F1F1F1} \bfseries 0 & {\cellcolor[HTML]{006837}} \color[HTML]{F1F1F1} 99.42 ($\pm$0.03) \\
 &  & \color[HTML]{cfcfcf} \xmark & \cmark & \color[HTML]{cfcfcf} \xmark & C & {\cellcolor[HTML]{0A7B41}} \color[HTML]{F1F1F1} 0.04 ($\pm$0.02) & {\cellcolor[HTML]{006837}} \color[HTML]{F1F1F1} \bfseries 0 & {\cellcolor[HTML]{006837}} \color[HTML]{F1F1F1} \bfseries 99.54 ($\pm$0.02) \\
 &  & \cmark & \color[HTML]{cfcfcf} \xmark & \color[HTML]{cfcfcf} \xmark & \color[HTML]{cfcfcf} \xmark & {\cellcolor[HTML]{FEE28F}} \color[HTML]{000000} \itshape 0.59 ($\pm$0.10) & {\cellcolor[HTML]{F26841}} \color[HTML]{F1F1F1} \itshape 80.47 & {\cellcolor[HTML]{006837}} \color[HTML]{F1F1F1} 99.44 ($\pm$0.01) \\
 &  & \cmark & \cmark & \color[HTML]{cfcfcf} \xmark & P & {\cellcolor[HTML]{30A356}} \color[HTML]{F1F1F1} 0.13 ($\pm$0.07) & {\cellcolor[HTML]{006837}} \color[HTML]{F1F1F1} \bfseries 0 & {\cellcolor[HTML]{006837}} \color[HTML]{F1F1F1} 99.46 ($\pm$0.10) \\
 &  & \cmark & \cmark & \color[HTML]{cfcfcf} \xmark & C & {\cellcolor[HTML]{07753E}} \color[HTML]{F1F1F1} \bfseries 0.03 ($\pm$0.02) & {\cellcolor[HTML]{006837}} \color[HTML]{F1F1F1} \bfseries 0 & {\cellcolor[HTML]{006837}} \color[HTML]{F1F1F1} 99.52 ($\pm$0.02) \\
\cline{2-9}
 & \multirow[c]{9}{*}{\ac{ViT}} & \color[HTML]{cfcfcf} \xmark & \color[HTML]{cfcfcf} \xmark & \color[HTML]{cfcfcf} \xmark & \color[HTML]{cfcfcf} \xmark & {\cellcolor[HTML]{A90426}} \color[HTML]{F1F1F1} 0.99 ($\pm$0.00) & {\cellcolor[HTML]{A50026}} \color[HTML]{F1F1F1} \itshape 100 & {\cellcolor[HTML]{006837}} \color[HTML]{F1F1F1} 98.84 ($\pm$0.00) \\
 &  & \color[HTML]{cfcfcf} \xmark & \color[HTML]{cfcfcf} \xmark & \color[HTML]{cfcfcf} \xmark & P & {\cellcolor[HTML]{C21C27}} \color[HTML]{F1F1F1} 0.94 ($\pm$0.03) & {\cellcolor[HTML]{A50026}} \color[HTML]{F1F1F1} \itshape 100 & {\cellcolor[HTML]{016A38}} \color[HTML]{F1F1F1} 98.63 ($\pm$0.00) \\
 &  & \color[HTML]{cfcfcf} \xmark & \color[HTML]{cfcfcf} \xmark & \color[HTML]{cfcfcf} \xmark & C & {\cellcolor[HTML]{A50026}} \color[HTML]{F1F1F1} \itshape 1.00 ($\pm$0.00) & {\cellcolor[HTML]{A50026}} \color[HTML]{F1F1F1} \itshape 100 & {\cellcolor[HTML]{006837}} \color[HTML]{F1F1F1} 98.95 ($\pm$0.03) \\
 &  & \color[HTML]{cfcfcf} \xmark & \cmark & \color[HTML]{cfcfcf} \xmark & \color[HTML]{cfcfcf} \xmark & {\cellcolor[HTML]{AF0926}} \color[HTML]{F1F1F1} 0.98 ($\pm$0.01) & {\cellcolor[HTML]{A50026}} \color[HTML]{F1F1F1} \itshape 100 & {\cellcolor[HTML]{006837}} \color[HTML]{F1F1F1} 98.85 ($\pm$0.00) \\
 &  & \color[HTML]{cfcfcf} \xmark & \cmark & \color[HTML]{cfcfcf} \xmark & P & {\cellcolor[HTML]{05713C}} \color[HTML]{F1F1F1} 0.02 ($\pm$0.04) & {\cellcolor[HTML]{006837}} \color[HTML]{F1F1F1} \bfseries 0 & {\cellcolor[HTML]{026C39}} \color[HTML]{F1F1F1} \itshape 98.36 ($\pm$0.00) \\
 &  & \color[HTML]{cfcfcf} \xmark & \cmark & \color[HTML]{cfcfcf} \xmark & C & {\cellcolor[HTML]{6BBF64}} \color[HTML]{000000} 0.21 ($\pm$0.06) & {\cellcolor[HTML]{006837}} \color[HTML]{F1F1F1} \bfseries 0 & {\cellcolor[HTML]{006837}} \color[HTML]{F1F1F1} 99.04 ($\pm$0.00) \\
 &  & \cmark & \color[HTML]{cfcfcf} \xmark & \color[HTML]{cfcfcf} \xmark & \color[HTML]{cfcfcf} \xmark & {\cellcolor[HTML]{AF0926}} \color[HTML]{F1F1F1} 0.98 ($\pm$0.02) & {\cellcolor[HTML]{A50026}} \color[HTML]{F1F1F1} \itshape 100 & {\cellcolor[HTML]{006837}} \color[HTML]{F1F1F1} \bfseries 99.09 ($\pm$0.00) \\
 &  & \cmark & \cmark & \color[HTML]{cfcfcf} \xmark & P & {\cellcolor[HTML]{006837}} \color[HTML]{F1F1F1} \bfseries 0.00 ($\pm$0.04) & {\cellcolor[HTML]{006837}} \color[HTML]{F1F1F1} \bfseries 0 & {\cellcolor[HTML]{006837}} \color[HTML]{F1F1F1} 98.81 ($\pm$0.00) \\
 &  & \cmark & \cmark & \color[HTML]{cfcfcf} \xmark & C & {\cellcolor[HTML]{30A356}} \color[HTML]{F1F1F1} 0.13 ($\pm$0.04) & {\cellcolor[HTML]{006837}} \color[HTML]{F1F1F1} \bfseries 0 & {\cellcolor[HTML]{006837}} \color[HTML]{F1F1F1} 98.91 ($\pm$0.00) \\

 \midrule
 
 \multirow[c]{18}{*}{\rotatebox{90}{CIFAR-10}} & \multirow[c]{9}{*}{ResNet-18} & \color[HTML]{cfcfcf} \xmark & \color[HTML]{cfcfcf} \xmark & \color[HTML]{cfcfcf} \xmark & \color[HTML]{cfcfcf} \xmark & {\cellcolor[HTML]{FFF0A6}} \color[HTML]{000000} 0.55 ($\pm$0.11) & {\cellcolor[HTML]{FA9B58}} \color[HTML]{000000} 72.66 & {\cellcolor[HTML]{036E3A}} \color[HTML]{F1F1F1} 73.99 ($\pm$0.13) \\
 &  & \color[HTML]{cfcfcf} \xmark & \color[HTML]{cfcfcf} \xmark & \color[HTML]{cfcfcf} \xmark & P & {\cellcolor[HTML]{98D368}} \color[HTML]{000000} 0.28 ($\pm$0.20) & {\cellcolor[HTML]{42AC5A}} \color[HTML]{F1F1F1} 15.62 & {\cellcolor[HTML]{006837}} \color[HTML]{F1F1F1} \bfseries 74.82 ($\pm$0.27) \\
 &  & \color[HTML]{cfcfcf} \xmark & \color[HTML]{cfcfcf} \xmark & \color[HTML]{cfcfcf} \xmark & C & {\cellcolor[HTML]{CFEB85}} \color[HTML]{000000} 0.38 ($\pm$0.14) & {\cellcolor[HTML]{66BD63}} \color[HTML]{F1F1F1} 20.31 & {\cellcolor[HTML]{026C39}} \color[HTML]{F1F1F1} 74.23 ($\pm$0.57) \\
 &  & \color[HTML]{cfcfcf} \xmark & \cmark & \color[HTML]{cfcfcf} \xmark & \color[HTML]{cfcfcf} \xmark & {\cellcolor[HTML]{FFF0A6}} \color[HTML]{000000} 0.55 ($\pm$0.11) & {\cellcolor[HTML]{FCA85E}} \color[HTML]{000000} 71.09 & {\cellcolor[HTML]{0B7D42}} \color[HTML]{F1F1F1} 71.98 ($\pm$0.20) \\
 &  & \color[HTML]{cfcfcf} \xmark & \cmark & \color[HTML]{cfcfcf} \xmark & P & {\cellcolor[HTML]{6BBF64}} \color[HTML]{000000} 0.21 ($\pm$0.11) & {\cellcolor[HTML]{006837}} \color[HTML]{F1F1F1} \bfseries 0 & {\cellcolor[HTML]{06733D}} \color[HTML]{F1F1F1} 73.25 ($\pm$0.06) \\
 &  & \color[HTML]{cfcfcf} \xmark & \cmark & \color[HTML]{cfcfcf} \xmark & C & {\cellcolor[HTML]{0F8446}} \color[HTML]{F1F1F1} \bfseries 0.06 ($\pm$0.03) & {\cellcolor[HTML]{006837}} \color[HTML]{F1F1F1} \bfseries 0 & {\cellcolor[HTML]{0C7F43}} \color[HTML]{F1F1F1} \itshape 71.55 ($\pm$0.23) \\
 &  & \cmark & \color[HTML]{cfcfcf} \xmark & \color[HTML]{cfcfcf} \xmark & \color[HTML]{cfcfcf} \xmark & {\cellcolor[HTML]{FED07E}} \color[HTML]{000000} \itshape 0.63 ($\pm$0.10) & {\cellcolor[HTML]{D83128}} \color[HTML]{F1F1F1} \itshape 89.84 & {\cellcolor[HTML]{036E3A}} \color[HTML]{F1F1F1} 74.03 ($\pm$0.13) \\
 &  & \cmark & \cmark & \color[HTML]{cfcfcf} \xmark & P & {\cellcolor[HTML]{30A356}} \color[HTML]{F1F1F1} 0.13 ($\pm$0.08) & {\cellcolor[HTML]{006837}} \color[HTML]{F1F1F1} \bfseries 0 & {\cellcolor[HTML]{06733D}} \color[HTML]{F1F1F1} 73.27 ($\pm$0.42) \\
 &  & \cmark & \cmark & \color[HTML]{cfcfcf} \xmark & C & {\cellcolor[HTML]{0F8446}} \color[HTML]{F1F1F1} \bfseries 0.06 ($\pm$0.03) & {\cellcolor[HTML]{006837}} \color[HTML]{F1F1F1} \bfseries 0 & {\cellcolor[HTML]{0B7D42}} \color[HTML]{F1F1F1} 72.02 ($\pm$0.28) \\
\cline{2-9}
 & \multirow[c]{9}{*}{\ac{ViT}} & \color[HTML]{cfcfcf} \xmark & \color[HTML]{cfcfcf} \xmark & \color[HTML]{cfcfcf} \xmark & \color[HTML]{cfcfcf} \xmark & {\cellcolor[HTML]{D62F27}} \color[HTML]{F1F1F1} 0.90 ($\pm$0.05) & {\cellcolor[HTML]{A50026}} \color[HTML]{F1F1F1} \itshape 100 & {\cellcolor[HTML]{18954F}} \color[HTML]{F1F1F1} 63.71 ($\pm$0.00) \\
 &  & \color[HTML]{cfcfcf} \xmark & \color[HTML]{cfcfcf} \xmark & \color[HTML]{cfcfcf} \xmark & P & {\cellcolor[HTML]{F16640}} \color[HTML]{F1F1F1} 0.81 ($\pm$0.07) & {\cellcolor[HTML]{A50026}} \color[HTML]{F1F1F1} \itshape 100 & {\cellcolor[HTML]{15904C}} \color[HTML]{F1F1F1} 64.53 ($\pm$0.00) \\
 &  & \color[HTML]{cfcfcf} \xmark & \color[HTML]{cfcfcf} \xmark & \color[HTML]{cfcfcf} \xmark & C & {\cellcolor[HTML]{A50026}} \color[HTML]{F1F1F1} \itshape 1.00 ($\pm$0.00) & {\cellcolor[HTML]{A50026}} \color[HTML]{F1F1F1} \itshape 100 & {\cellcolor[HTML]{199750}} \color[HTML]{F1F1F1} 63.44 ($\pm$0.38) \\
 &  & \color[HTML]{cfcfcf} \xmark & \cmark & \color[HTML]{cfcfcf} \xmark & \color[HTML]{cfcfcf} \xmark & {\cellcolor[HTML]{E24731}} \color[HTML]{F1F1F1} 0.86 ($\pm$0.05) & {\cellcolor[HTML]{A50026}} \color[HTML]{F1F1F1} \itshape 100 & {\cellcolor[HTML]{2DA155}} \color[HTML]{F1F1F1} 61.83 ($\pm$0.43) \\
 &  & \color[HTML]{cfcfcf} \xmark & \cmark & \color[HTML]{cfcfcf} \xmark & P & {\cellcolor[HTML]{026C39}} \color[HTML]{F1F1F1} \bfseries 0.01 ($\pm$0.02) & {\cellcolor[HTML]{006837}} \color[HTML]{F1F1F1} \bfseries 0 & {\cellcolor[HTML]{2DA155}} \color[HTML]{F1F1F1} 61.99 ($\pm$0.00) \\
 &  & \color[HTML]{cfcfcf} \xmark & \cmark & \color[HTML]{cfcfcf} \xmark & C & {\cellcolor[HTML]{219C52}} \color[HTML]{F1F1F1} 0.11 ($\pm$0.05) & {\cellcolor[HTML]{006837}} \color[HTML]{F1F1F1} \bfseries 0 & {\cellcolor[HTML]{36A657}} \color[HTML]{F1F1F1} 61.31 ($\pm$0.00) \\
 &  & \cmark & \color[HTML]{cfcfcf} \xmark & \color[HTML]{cfcfcf} \xmark & \color[HTML]{cfcfcf} \xmark & {\cellcolor[HTML]{CC2627}} \color[HTML]{F1F1F1} 0.92 ($\pm$0.04) & {\cellcolor[HTML]{A50026}} \color[HTML]{F1F1F1} \itshape 100 & {\cellcolor[HTML]{006837}} \color[HTML]{F1F1F1} \bfseries 69.47 ($\pm$0.00) \\
 &  & \cmark & \cmark & \color[HTML]{cfcfcf} \xmark & P & {\cellcolor[HTML]{026C39}} \color[HTML]{F1F1F1} \bfseries 0.01 ($\pm$0.01) & {\cellcolor[HTML]{006837}} \color[HTML]{F1F1F1} \bfseries 0 & {\cellcolor[HTML]{006837}} \color[HTML]{F1F1F1} 69.30 ($\pm$0.00) \\
 &  & \cmark & \cmark & \color[HTML]{cfcfcf} \xmark & C & {\cellcolor[HTML]{0F8446}} \color[HTML]{F1F1F1} 0.06 ($\pm$0.03) & {\cellcolor[HTML]{006837}} \color[HTML]{F1F1F1} \bfseries 0 & {\cellcolor[HTML]{36A657}} \color[HTML]{F1F1F1} \itshape 61.18 ($\pm$0.00) \\
\bottomrule
\end{tabular}%
}
\end{table*}

The results for these experiments are presented in \autoref{tab:results_xcomb_models}.
The tendencies for model utility and privacy protection are consistent with the observations made for the \ac{CNN}. 
For the ResNet-18, only using a \ac{PM} (either \ac{PRECODE} or \ac{CVB}) already reduces \ac{ASR} to below $21\%$ for both datasets. 
\ac{DP} alone provides a slight reduction in \ac{ASR}, by $14.85\%$ on MNIST and $1.57\%$ on CIFAR-10
For the \ac{ViT}, these defenses are insufficient as \ac{ASR} remains at $100\%$ if only single defenses are applied.
For both models and datasets, a combination of \ac{DP}+\ac{PRECODE} or \ac{DP}+\ac{CVB} is required to reduce \ac{ASR} to $0\%$. 
However, there is no defense combination that consistently delivers the best trade-off between model utility and privacy across all settings.
On MNIST, the ResNet-18 and \ac{ViT} achieve the best trade-off with \ac{DP}+\ac{CVB}, reducing \ac{ASR} to $0\%$ and increasing accuracy by $0.03\%$ and $0.20\%$ compared to the unprotected baseline models, respectively. 
In contrast, on CIFAR-10, \ac{DP}+\ac{CVB} reduces model utility compared to the unprotected baseline. 
In this case, the best trade-off is achieved with the DO+\ac{DP}+\ac{PRECODE} combination, which reduces \ac{ASR} to $0\%$ for both models.
For the ResNet-18, this combination leads to a slight decrease in model utility by $0.72\%$, while for the \ac{ViT}, it results in a substantial accuracy increase of $5.59\%$ compared to the baseline model. 
This improvement is likely due to the positive effect of Dropout on the \ac{ViT} architecture, which has also been observed in~\parencite{scheliga2023dropout}.

\FloatBarrier
\section{Conclusion}
\label{sec:conclusion}
Privacy preservation in \ac{FL} is a rapidly evolving field of research that continuously introduces new attacks and enhanced defense mechanisms. 
First, we adopted an attacker’s perspective to test the limits of commonly applied defense mechanisms. 
Our findings show that attackers can bypass these defenses by sufficiently approximating and mimicking the client's stochastic gradient computation process. 
We implemented three novel targeted attacks and demonstrated that the protection promised by \ac{GP}, \ac{PRECODE}, and \ac{CVB} can be significantly weakened. 
Ultimately, we found that no individually applied defense mechanism provides adequate privacy protection.

To address this issue, we proposed to combine multiple defenses. 
Since each defense alters gradient computation in a different way, it becomes increasingly challenging for attackers to fully mimic the gradient computation process and conduct a successful \ac{GI} attack. 
We conducted a comprehensive ablation study to evaluate various defense combinations and their impact on both model utility and privacy. 
To establish an upper bound for privacy leakage, we developed \ac{CDIA}, an attack specifically targeting each defense applied by the client during local training.
Across various scenarios, we found that combining \ac{DP} with either \ac{PRECODE} or \ac{CVB} is essential to protect client privacy, \ie reducing the \ac{ASR} to 0\%. 
These combinations also frequently resulted in the highest model utility compared to other defense combinations, single defenses, and even the unprotected baseline.
This paper highlights the importance to thoroughly evaluate and analyze defense mechanisms that are supposed to protect client privacy in \ac{FL}. 
To this end, we have developed new strategies to more effectively resist privacy threats in collaborative environments.

\section*{Funding Details}
This work was supported by the Thuringian Ministry of Economics, Science and Digital Society under Grant 5575/10-3.

\section*{Disclosure Statement}
This paper does not have potential conflict of interest.

\section*{Data Availability Statement}
All dataset that were used for the experiments are publicly available: MNIST~\parencite{lecun1998gradient}; CIFAR-10~\parencite{krizhevsky2009learning}.

\printbibliography

\end{document}